\documentclass{article}

\PassOptionsToPackage{numbers, compress}{natbib}
\usepackage[preprint]{neurips_2026}

\usepackage[utf8]{inputenc} 
\usepackage[T1]{fontenc}    
\usepackage{hyperref}       
\usepackage{url}            
\usepackage{booktabs}       
\usepackage{amsfonts}       
\usepackage{nicefrac}       
\usepackage{microtype}      
\usepackage{xcolor}         
\usepackage{graphicx}
\usepackage{subcaption}
\usepackage[most]{tcolorbox}
\usepackage{multirow}
\usepackage[table]{xcolor}
\usepackage{wrapfig}
\usepackage{float}
\usepackage{titletoc}

\usepackage{enumitem}

\newcommand{\stitle}[1]{\vspace*{0.3em}\noindent{\bf #1.\/}}

\title{CAM: Question Answering on Entity-Centric Videos with Continuous Extraction and Adaptive Querying}

\author{%
  Yizhou Tian$^{1}$ \quad
  Zizhe Chen$^{1}$ \quad
  Shiyuan Deng$^{2}$ \quad
  Garry Yang$^{1}$ \quad
  Zijie Dai$^{1}$ \\
  \bfseries Luohao Pan$^{1}$ \quad
  Hao Lin$^{1}$ \quad
  Peiqi Yin$^{1}$ \quad
  Xiao Yan$^{3}$ \quad
  James Cheng$^{1}$ \\[1.2ex]
  \normalfont $^{1}$Department of Computer Science and Engineering, The Chinese University of Hong Kong \\
  $^{2}$Knowin AI \qquad
  $^{3}$Institute for Math and AI, Wuhan University
}

\begin{document}

\maketitle

\begin{abstract}

Memory facilitates question answering over long videos by extracting and retrieving facts to fit within the limited context windows of multimodal LLMs (MLLMs).
Existing solutions typically extract independent memory entries from fixed-length video clips and thus cannot capture high-level semantics that need to be summarized over extended time periods, such as character traits and relations.
Moreover, they rely solely on similarity-based retrieval and may fail to retrieve the fine-grained details required for question answering. To tackle these problems, we propose CAM, featuring \textit{continuous extraction} for high-level semantics and \textit{adaptive querying} for fine-grained details. In particular, CAM stores the entities and relations extracted from video clips in a knowledge graph. To capture the high-level semantics of each entity or relation, CAM summarizes the local subgraph of the target entity or relation once the subgraph reaches a predefined size. To retrieve the fine-grained details required for question answering, CAM supports multiple search methods, including knowledge graph traversal, video re-watching, and audio listening. It utilizes a planner-executor-verifier pipeline to adaptively compose these search methods according to question intent.
Evaluations on three benchmarks show that CAM outperforms SOTA baselines and improves their accuracy by up to 23 percentage points. Code is available at \url{https://github.com/Jake-Tian/CAM}.

\end{abstract}

\section{Introduction}
\label{sec:introduction}

The latest multimodal LLMs (MLLMs) can perceive and reason over complex visual and linguistic signals \cite{gpt5, gemini3, qwen3vl}. However, directly using MLLMs to process long videos is challenging since the context windows of MLLMs are limited, and it is costly to process many input tokens. Real-world videos, such as observing continuous robotic operations \cite{openvla}, streaming first-person footage from AI glasses \cite{egolife}, or following a plot in a cinematic movie \cite{lvbench}, are usually \textit{entity-centric} and record the attributes, events and relations for a persistent set of entities (e.g., characters and objects). As such, video memory systems~\cite{egolife,worldmm,m3agent} first extract facts about these entities from the videos and then retrieve only the related facts during question answering to reduce input tokens.

\begin{figure}
    \centering
    \includegraphics[width=1\linewidth]{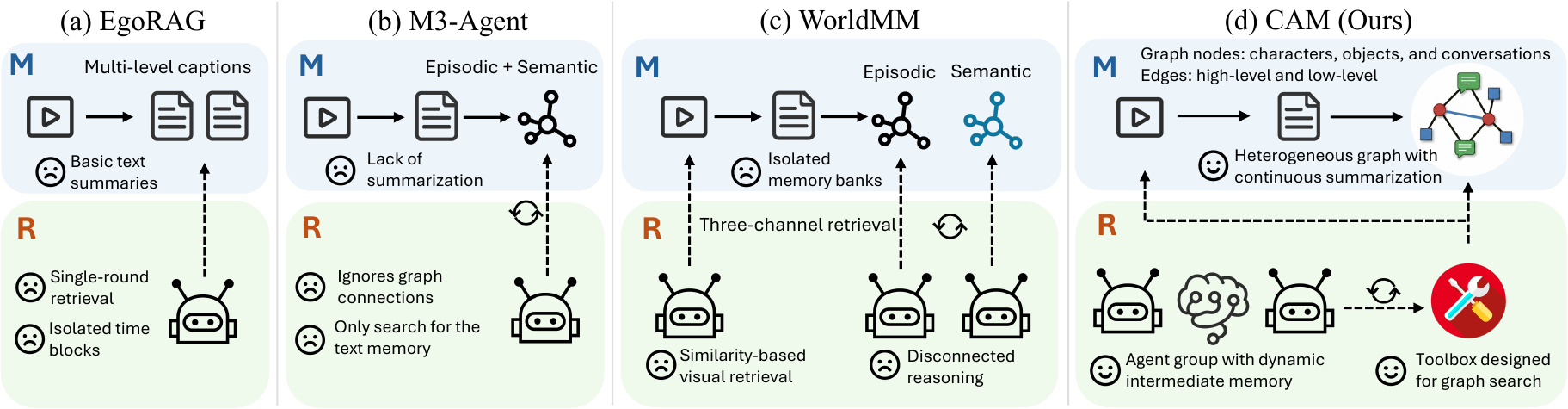}
    \caption{Comparing existing video memory systems and our CAM. The blue ``M" denotes memory construction, while yellow ``R" denotes question answering. (a) EgoRAG~\cite{egolife} generates a text caption for each video clip and searches these captions. (b) M3-Agent~\cite{m3agent} extracts facts related to each entity. (c)WorldMM~\cite{worldmm} builds multiple memory modules but completely decouples visual features from textual knowledge, leading to fragmented retrieval. (d) CAM (Ours) organizes the entities and relations into a temporally grounded, heterogeneous knowledge graph, continuously summarizes high-level semantics about each entity and relation, and searches the knowledge graph adaptively using multiple methods.}
    \label{fig:alpha}
\end{figure}

We illustrate the designs of three representative video memory systems in Figure~\ref{fig:alpha}. In particular, EgoRAG~\cite{egolife} segments a video into short clips and generates hierarchical text captions on rigid temporal boundaries. To answer questions, these captions are searched based on text similarity. M3-Agent~\cite{m3agent} explicitly models entities and organizes the facts according to their related entities. While it constructs an entity-centric memory, its retrieval mechanism fundamentally underutilizes the graph topology, relying primarily on flat textual similarity matching that struggles to navigate complex relational chains. WorldMM~\cite{worldmm} attempts to capture both temporal events and semantics by maintaining separate episodic, semantic, and visual memory modules. Despite this, WorldMM physically segregates its knowledge into decoupled memory banks. Because visual features are isolated from the textual knowledge graph, the agent is forced to conduct independent, fragmented retrieval streams, failing to achieve a unified multimodal understanding of complex interactions.

We find that existing video memory systems struggle on two fronts: capturing high-level semantics (e.g., character traits, relationships, narrative structure) and retrieving fine-grained details (e.g., characters' behaviors and object spatial relations) — both of which are crucial for accurate question answering.
On the high-level side, semantics such as personality or social relations cannot be read off any single clip; they emerge only after accumulating many events over an extended time period. Existing systems, however, process each short clip independently and rarely consolidate evidence across clips. This omission is not benign: as a video grows longer, the raw facts associated with each entity grow rapidly, and without proper summarization, retrieval is forced either to overflow the MLLM's context window with low-level episodic facts or to prune them via truncation and risk dropping the very evidence needed to infer high-level semantics at query time.
On the fine-grained side, two issues compound. First, although some prior systems organize memory as a knowledge graph, retrieval still falls back to flat textual similarity over node descriptions, leaving the graph topology — and the relational chains it encodes — largely unexploited. Second, memory construction is inherently lossy: subtle visual cues and acoustic details are inevitably discarded when clips are compressed into text, and which details matter can only be determined once the question is known. A retrieval mechanism that searches only the constructed memory therefore has no recourse when the answer hinges on a detail that was dropped during construction.

To tackle the limitations of existing video memory systems, we propose Continuous Adaptive Memory (CAM), which is illustrated in Figure~\ref{fig:alpha}(d). In particular, CAM uses a temporally grounded, heterogeneous knowledge graph to store the extracted entities and relations along with their associated facts. To capture high-level semantics, CAM introduces evidence-driven summarization. Rather than operating at fixed time intervals, summarization is treated as an asynchronous operation dictated by subgraph maturity. A localized history is summarized into high-level traits or relational edges only when a specific entity or interaction pair accumulates a sufficient density of incident edges or attributes. This ensures that the summaries are grounded on stable, diverse observations rather than transient anomalies, which improves memory quality.

Besides similarity search, CAM allows multiple additional search methods to retrieve the facts required by the target question, such as traversing the knowledge graph, re-watching a segment of the video, listening to the audio, and counting the frequencies of entities and events. These search methods are more flexible and can retrieve the fine-grained details that are lost during memory construction. To jointly utilize these search methods, CAM proposes an adaptive retrieval workflow, where a planner configures the queries for all search methods, an executor runs each search method, and a verifier checks if the retrieved contents can answer the target question. The plan-execute-verify workflow can also run iteratively for multiple rounds to pinpoint the contents to retrieve.


To evaluate CAM, we conduct experiments on three long video benchmarks and compare with four SOTA video memory systems along with strong base models and online services. The results show that CAM consistently outperforms all baselines in question answering accuracy, and compared with the best-performing baseline, CAM improves accuracy by 12.93 percentage points on average and up to 23.1 percentage points. The ablation study shows that our designs are effective in improving accuracy, and profiling shows that our token cost is comparable to the baselines.    

To summarize, we make the following contributions in this paper.
\begin{itemize}[leftmargin=*, topsep=0pt, itemsep=0pt, parsep=0pt]
    \item We identify the challenges of designing memory for long videos (i.e., capturing high-level semantics and retrieving low-level details) and analyze the limitations of existing video memory systems.
    \item To capture high-level semantics, we model the entities or relations extracted from video clips using a heterogeneous knowledge graph and summarize each entity and relation once the local subgraph of the target entity and relation reaches a sufficient size.
    \item To retrieve the fine-grained details required by each question, we propose a rich set of search methods and utilize these search methods adaptively with an iterative workflow.
\end{itemize}

\section{Preliminary and Related Work}
We study complex question answering over entity-centric long videos, where recurring entities, such as humans, robots, objects, or conversational participants, persist and interact across clips. This setting requires maintaining entity continuity, tracking evolving relations, and answering questions from fine-grained episodic evidence and emergent high-level semantics. Because an entire long video cannot fit into the context window of an MLLM, the key challenge is to build a persistent memory that preserves multimodal details while supporting long-horizon memorization and retrieval.

Long video understanding methods commonly use memory-centric designs or retrieval-augmented generation to overcome context limits~\cite{moviechat,luo2024videorag}. Prior systems rely on flat summaries~\cite{m3agent,egolife}, sequential episodic memories~\cite{hippomm}, or fixed-interval updates, which are less suitable for entity-centric videos whose semantics emerge asynchronously from accumulated interactions.
Graph-based methods introduce more structured representations~\cite{chu2025entitygraph,VideoRAG,malik2025ravu}, but often use graph topology mainly as an auxiliary retrieval index.
CAM instead builds a temporally grounded heterogeneous semantic graph, abstracts mature local subgraphs, and uses a tool-augmented agent for graph traversal, targeted retrieval, and video rewatching.
A comprehensive review is provided in Appendix~\ref{sec:appendix_related_work}.
\section{Methodology}

\begin{figure*}[!t]
    \centering
    \includegraphics[width=1\linewidth]{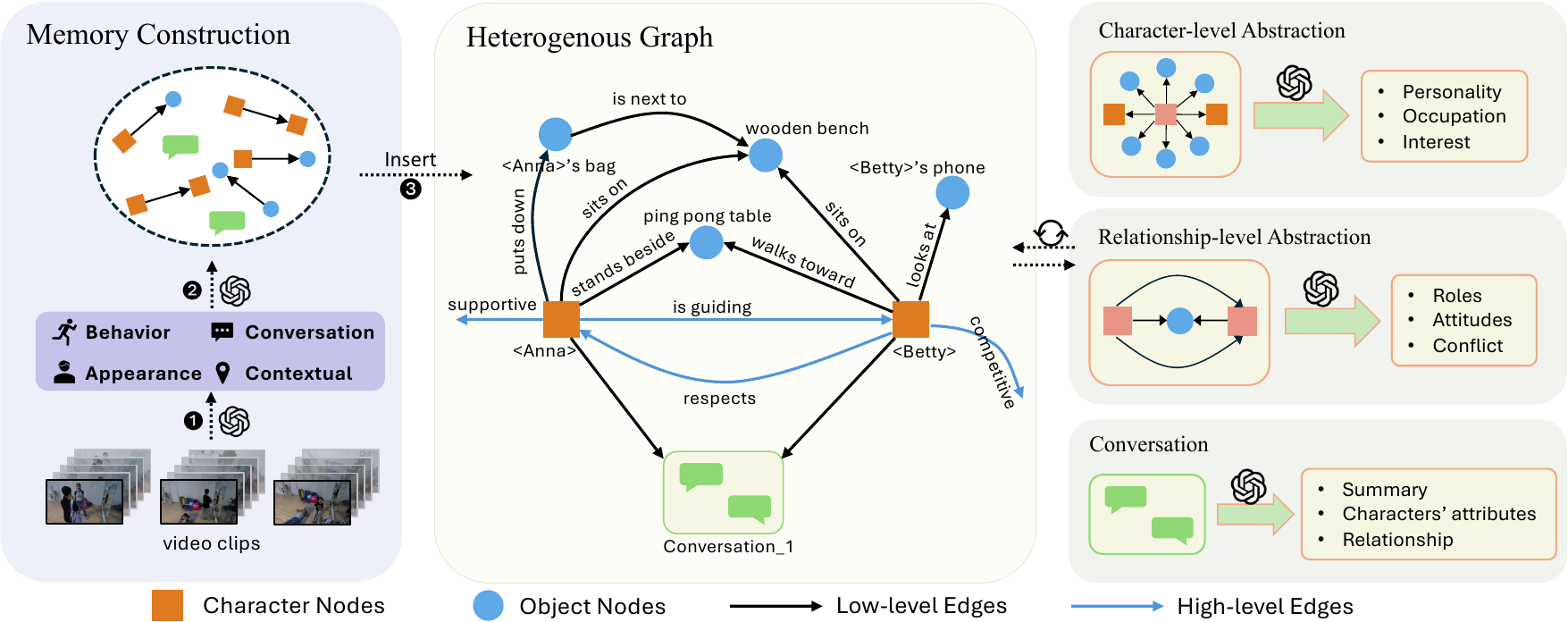}
    \caption{The memory construction workflow of CAM. This unified architecture effectively bridges the gap between low-level episodic observations and emergent high-level semantics, supporting robust and efficient question answering over extended temporal horizons.}
    \label{fig:memory_construction}
\end{figure*}

Our framework, Continuous Adaptive Memory (CAM), enhances long-form video question answering by progressively transforming clip-level observations into a temporally grounded, heterogeneous relational graph to drive complex, multi-hop agent reasoning. As illustrated in Figure~\ref{fig:memory_construction}, a long video is first segmented into short clips, each processed by an MLLM to extract fine-grained episodic information, including character behaviors, conversations, and appearance descriptors. These observations are aligned across time and inserted into the memory graph as temporally annotated relational edges. Crucially, to capture high-level semantics that emerge across multiple observations, CAM replaces rigid, fixed-interval summarization with evidence-driven abstraction; it asynchronously synthesizes localized subgraphs into persistent character traits and social relationships only upon reaching structural maturity. At query time, CAM bypasses standard single-pass retrieval by deploying a tool-augmented agent workflow centered on a dynamically updating intermediate memory. Governed by a unified LLM planner, the agent systematically exploits the memory topology through structure-aware retrieval for relational deduction. When textual graph evidence is sparse or ambiguous, the agent dynamically invokes multimodal fallbacks—such as targeted video rewatching and audio sampling—to ground its reasoning in raw sensory details. This unified architecture effectively bridges the gap between low-level episodic observations and emergent high-level semantics, supporting robust and efficient question answering over extended temporal horizons.

\subsection{Memorization}
\label{subsec:memorization}

\subsubsection{Memory Graph Construction}

\stitle{Clip-level observation} 
The input video is segmented into fixed-length clips $\{c_1,\ldots, c_n\}$, each treated as an episodic observation unit. To preserve temporal continuity and enable consistent entity tracking, each clip $c_i$ is processed together with appearance descriptors from the preceding clip $A_{i-1}$, providing visual cues for cross-clip entity association. Given these inputs, an MLLM extracts complementary information from each clip:

\begin{equation*}
\bigl(B_i,\; S_i,\; A_i,\; K_i\bigr) \;\leftarrow\; \text{MLLM}\bigl(c_i,\; A_{i-1}\bigr).
\end{equation*}

Here, $B_i$ captures observable character behaviors, $S_i$ records dialogue transcripts, $A_i$ encodes appearance-based descriptors, and $K_i$ stores auxiliary context for later reasoning. Using $A_i$, characters are aligned across clips to ensure consistent identities over time, even under viewpoint changes, partial occlusions, or temporal gaps. After alignment, the behavior descriptions $B_i$ are transformed by an LLM into structured relational triples $(\textit{source},\; \textit{relation},\; \textit{target})$. Each triple represents an observed interaction and serves as a minimal, interpretable unit of episodic evidence.

\stitle{Heterogeneous memory graph integration}
All extracted signals are integrated into a heterogeneous memory graph
\begin{equation*}
G = (V, E),
\qquad
V = V_{\text{char}} \cup V_{\text{obj}} \cup V_{\text{conv}},
\end{equation*}

which serves as the core long-term video representation. Character and object nodes correspond to aligned entities across clips, while each transcript $S_i$ is represented as a conversation node linked to its speakers, enabling joint modeling of physical actions and dialogue with distinct semantic roles.

Relational triples extracted from the $i$-th clip are inserted into the graph as temporally annotated edges:
\[
e_{ij}
=
\bigl(
\textit{source},\;
\textit{relation}_j,\;
\textit{target},\;
t = i,\;
K_i
\bigr),
\]
where $\textit{relation}_j$ is the $j$-th relation extracted from $c_i$. Temporal indices and contextual metadata preserve the video’s chronological structure, supporting time-aware retrieval, causal reasoning, and episodic reconstruction. Overall, the memorization stage converts a long video into a structured, time-anchored semantic graph that underpins later abstraction and reasoning.

\subsubsection{Abstract Information Extraction}

Initial MLLM transcriptions inherently capture only explicitly visible, low-level episodic details within isolated frames. However, higher-order semantics, such as a character's intrinsic personality traits or social relationships, are not directly observable; they emerge over time through a continuous history of actions and interactions. Because video evidence accumulates unevenly (some characters stabilize early, while others remain sparse), abstracting this information at fixed temporal intervals inevitably yields premature or delayed summaries that fracture memory consistency. 

To bridge this semantic gap, CAM employs \emph{evidence-driven abstraction}. Rather than relying on rigid time steps, abstraction is treated as an asynchronous operation governed by subgraph maturity. CAM continuously monitors the structural density of the memory graph; summarization is triggered only when a specific entity or pair accumulates a sufficiently rich history of incident edges. This ensures that abstractions are strictly grounded in diverse, stable observations rather than transient anomalies. More details about evidence-driven abstraction are provided in Appendix~\ref{subsec:appendix_abstract}. 

When triggered, an LLM synthesizes the accumulated low-level history into a high-level semantic node or edge. This abstraction is injected back into the graph, while the original episodic evidence is preserved for future refinement. We apply this mechanism at two complementary levels:

\vspace{1mm}
\stitle{Character-level abstraction} 
Models individual behavioral tendencies. For each character $v \in V_{\text{char}}$, the system tracks the incrementally growing set of all incident episodic evidence, including actions and contextual states: $\mathcal{E}(v) = \{ e \in E \mid e.\text{source} = v \ \text{or} \ e.\text{target} = v \}.$ Once this specific subset reaches structural maturity, it is synthesized into an abstract character node.

\vspace{1mm}
\stitle{Relationship-level abstraction}
Captures persistent inter-character social dynamics. For any character pair $(u,v) \in V_{\text{char}} \times V_{\text{char}}$, the relevant evidence set encompasses both explicit interactions and implicit co-occurrences (e.g., sharing a location within a short temporal window): $\mathcal{E}(u,v) = \mathcal{E}_{\text{direct}}(u,v) \cup \mathcal{E}_{\text{indirect}}(u,v)$. Upon sufficient accumulation, this localized history is abstracted into a persistent relational edge between $u$ and $v$.

\subsection{Adaptive Reasoning}

\begin{figure*}[!t] 
    \centering
    \includegraphics[width=1\linewidth]{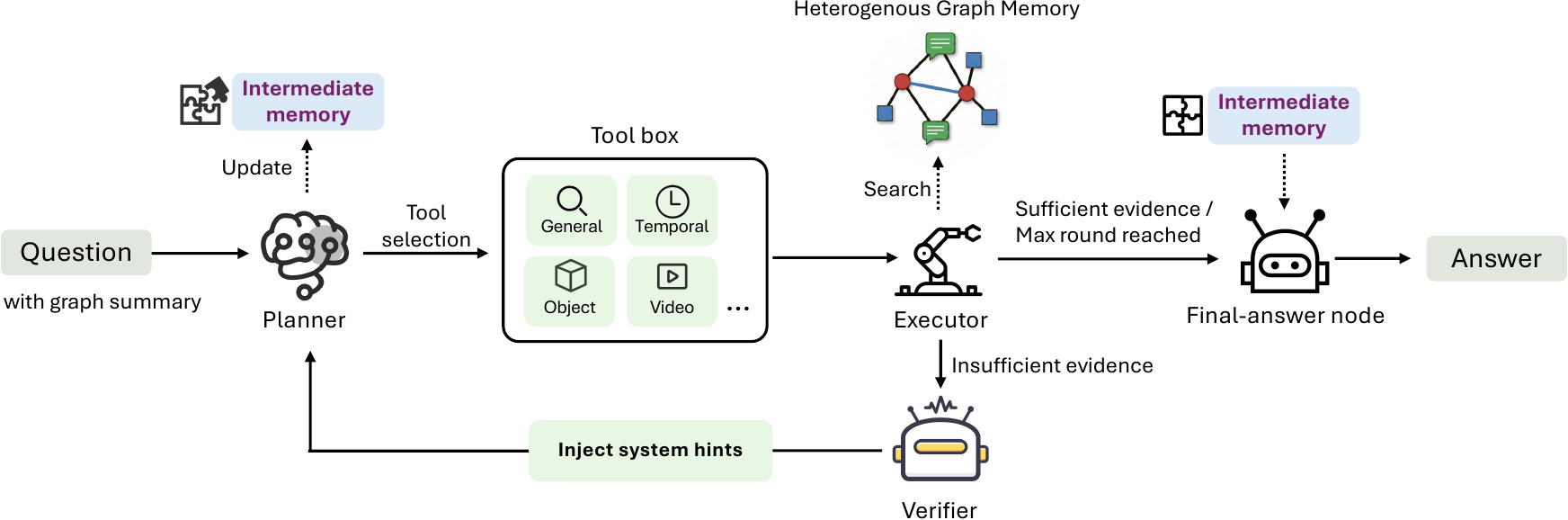}
    \caption{The adaptive reasoning workflow of CAM. This iterative feedback loop enables the system to dynamically refine its search strategy based on intermediate findings, ensuring that complex multi-step queries are resolved with high precision and minimal retrieval overhead.}
    \label{fig:adaptive_reasoning}
\end{figure*}

\subsubsection{Reasoning Workflow}

To effectively navigate the structured memory graph $\mathcal{G}$ and answer complex, multi-hop queries, we formulate the reasoning process as an iterative, tool-augmented state machine implemented in LangGraph, as depicted in Figure~\ref{fig:adaptive_reasoning}. Let the agent's state at turn $t$ be denoted as $\mathcal{S}_t = (\mathcal{Q}, \mathcal{C}_{meta}, \mathcal{M}_t, \mathcal{H}_t)$. Here, $\mathcal{Q}$ represents the user query, and $\mathcal{H}_t = \{e_1, e_2, \dots, e_{t-1}\}$ is the history of raw evidence gathered up to turn $t$. 

Crucially, $\mathcal{C}_{meta}$ represents the global graph metadata provided to the agent at initialization ($t=0$). This gives the agent a macroscopic overview of the context before execution, encompassing the video timespan, a list of main characters, the total number of extracted knowledge triples, and available conversations. Furthermore, $\mathcal{M}_t$ denotes a dynamically updated intermediate memory that synthesizes findings across turns.

The agent systematically gathers evidence through a controlled loop of planning, execution, and self-correction:

\begin{enumerate}[leftmargin=*, topsep=0pt, itemsep=0pt, partopsep=0pt]
    \item \textbf{Planner Node (LLM):} Acting as the central controller, the LLM analyzes the comprehensive current state $\mathcal{S}_t$, including the macro-level $\mathcal{C}_{meta}$ and the progressive deductions in $\mathcal{M}_t$, to determine the optimal next action $a_t \in \mathcal{A}$ and its corresponding arguments:
    \begin{equation}
        a_t, \text{args}_t = \text{LLM}(\mathcal{S}_t)
    \end{equation}
    
    \item \textbf{Executor Node:} This module invokes the selected tool $a_t$ to interact with the environment (the memory graph $\mathcal{G}$ or raw video). The execution yields new empirical evidence $e_t$:
    \begin{equation}
        e_t = \text{Execute}(a_t, \text{args}_t, \mathcal{G})
    \end{equation}
    Following execution, the history is updated ($\mathcal{H}_{t+1} = \mathcal{H}_t \cup \{e_t\}$), and the intermediate memory $\mathcal{M}_{t+1}$ updates to logically assimilate $e_t$, accumulating a structured chain of thought that gradually approaches the final answer.
    
    \item \textbf{Task Completion \& Verifier Node:} A verification function $\mathcal{V}$ evaluates the intermediate memory $\mathcal{M}_{t+1}$ to determine if the accumulated context resolves the query:
    \begin{equation}
        \mathcal{V}(\mathcal{S}_{t+1}) \rightarrow \{\text{Complete}, \text{Incomplete}, \text{Stuck}\}
    \end{equation}
    If incomplete and the max round is not reached, the loop iterates. If the state is classified as ``Stuck'' (e.g., detecting cyclic tool calls where $a_t = a_{t-1}$), the Verifier injects system-level heuristic hints $h$ into the state before routing back to the Planner, dynamically regularizing the LLM's search trajectory.
    
    \item \textbf{Final Answer Node:} The iterative process terminates when $\mathcal{V}(\mathcal{S}_{t+1}) = \text{Complete}$ or the budget is exhausted. The final state $\mathcal{S}_T$ transitions to the final answer node, where the collected evidence is organized to deduce the final answer $\hat{y}$. For multiple-choice questions, the output is constrained to a single categorical letter. For open-ended generative questions, the node structures the final output as a concise, single-sentence short answer.
\end{enumerate}

\subsubsection{Toolbox}

To empower the agent to actively navigate the heterogeneous memory graph and ground its reasoning in raw multimodal signals, we define a comprehensive suite of specialized tools. 

The foundational mechanism is the \textbf{General Search } tool. To handle queries requiring both global relations and fine-grained details, its logic is divided into three stages:

\vspace{1mm}
\noindent\textit{1) Weighted Query Decomposition.} The Planner Node parses the query $q$ into a semantic triple $\mathcal{T}_q = \langle s_q, r_q, t_q \rangle$ with a normalized weight vector $\mathbf{W}_q$. It extracts auxiliary context $I_q$ and determines a channel-specific retrieval budget $\mathbf{K}_q$.

\vspace{1mm}
\noindent\textit{2) Triple Similarity Computation.} To evaluate relevance within the graph $\mathcal{G}$, we compute similarities in a dense embedding space. To account for reversed semantic roles (e.g., subject vs. object), we compute a bidirectional alignment score for the entities and integrate it with the direct semantic matching of the relation to calculate the overall structural similarity:
\begin{equation}
\label{eq:s_struct}
\begin{aligned}
S_{\text{struct}} = &\max \big( w_s \cdot \mathrm{sim}(s_q, s_c) + w_t \cdot \mathrm{sim}(t_q, t_c), \; w_s \cdot \mathrm{sim}(s_q, t_c) + w_t \cdot \mathrm{sim}(t_q, s_c) \big) \\
&+ w_r \cdot \mathrm{sim}(r_q, r_c).
\end{aligned}
\end{equation}

\vspace{1mm}
\noindent\textit{3) Multi-Channel Retrieval.} Retrieval is partitioned by semantic roles. High-level abstractions are retrieved via $S_{\text{struct}}$. Low-level episodic relations require spatio-temporal grounding, scoring candidates via $S_{\text{struct}} + \lambda \cdot sim(I_q, K)$. Conversational content relies on direct semantic matching against the query, expanding retrieved utterances to include chronological neighbors ($d_{i-1}, d_{i+1}$) to preserve discursive coherence. All retrieved evidence is merged to flexibly integrate abstract, episodic, and dialogue contexts.

\vspace{2mm}
Beyond this primary search function, the agent is equipped with six additional specialized tools to facilitate explicit topological traversal, targeted adaptive querying, and raw sensory verification:

\begin{itemize}[leftmargin=*, topsep=0pt, itemsep=0pt, partopsep=0pt]
    
    \item \textbf{Temporal Search:} constrains retrieval to a window (t\_begin, t\_end). 

    \item \textbf{Graph Traversal:} retrieves semantically filtered neighborhoods around an anchor entity for multi-hop deduction.

    \item \textbf{Search Object:} embedding-based entity alignment to handle vocabulary mismatch.

    \item \textbf{Frequency Count:} aggregates occurrence counts of actions/objects/events.
    
    \item \textbf{Video Rewatch:} Acts as a critical multimodal fallback by directly rewatching the video to recover subtle visual nuances and continuous motion dynamics lost during text compression.
    
    \item \textbf{Audio Listening:} samples raw audio for non-speech acoustic cues.
\end{itemize}

A more detailed explanation of each tool is provided in Appendix~\ref{subsec:tool}. 

\subsection{Discussion}
Despite its strong reasoning capability, CAM leaves room for further optimization. First, CAM is most suitable for entity-centric long videos, such as robotic operations or character-driven narratives, where persistent entities and evolving relations are important. For less entity-centric videos, such as lectures or advertisements, the benefits may be more limited. Second, since useful tools can vary across domains and question types, future work may explore dynamic tool selection to better balance robustness, accuracy, and inference efficiency.

\section{Experimental Evaluation}
\label{sec:experiment}




\subsection{Experiment Settings}
\label{subsec:setting}
\stitle{Datasets and Metrics}
We evaluate on three entity-centric long-video benchmarks. \textbf{M3-Bench-robot} (averaging 34 minutes, robotic views) and \textbf{M3-Bench-web} (27 minutes, YouTube) assess GPT-5.2-judged accuracy across five reasoning categories: multi-evidence (ME), multi-hop (MH), cross-modal (CM), person understanding (PU), and general knowledge (GK). We validate the judge against GPT-4o and human annotations in Appendix~\ref{sec:implementation}. Finally, \textbf{HippoVlog} features 25 videos (averaging 27.2 minutes) and is evaluated using 1,000 multiple-choice questions.

\stitle{Baselines and Implementation}
We compare CAM against a comprehensive suite of base video LLMs (GPT-5.2, Gemini-3-Pro, Qwen3-VL-flash), streaming models (MovieChat, MA-LLM, Flash-VStream), and memory-based frameworks (EgoRAG, HippoMM, M3-Agent, WorldMM). For fair comparison, CAM and all memory-based baselines, except HippoMM, which exclusively uses Qwen, utilize GPT-5.2 as the core backbone, while CAM additionally evaluates Qwen3-VL-flash for video and Qwen3.5-flash for reasoning variants. HippoMM is restricted to Qwen because its extensive preprocessing steps, such as dynamic segmentation, would make GPT-based processing computationally prohibitive and prevent the system from precessing videos faster than real time. Videos are sampled at 1 fps and processed in 30-second fixed intervals (dynamically segmented 5-10-second clips for HippoMM), with audio transcribed into subtitles for unified multimodal reasoning. Comprehensive implementation details are provided in Appendix~\ref{sec:implementation}.

\subsection{Main Results}
\begin{table*}[!t]
    \centering
        \caption{Question answering accuracy. All video memory methods use GPT as the MLLM except HippoMM due to efficiency considerations. Besides GPT, we also report the accuracy of our CAM when using Qwen as the MLLM to demonstrate generality.} 
    \renewcommand{\arraystretch}{1.2} 
    \definecolor{myblue}{RGB}{232, 242, 252}
    \newcommand{\cb}[1]{{\setlength{\fboxsep}{2pt}\colorbox{myblue}{\makebox[2em]{#1}}}}
    
    \setlength{\tabcolsep}{3pt} 
    
    \resizebox{0.9\linewidth}{!}{%
    \begin{tabular}{@{} l ccccc c ccccc c c @{}}
        \toprule
        \multirow{2}{*}{\textbf{Method}} & \multicolumn{6}{c}{\textbf{M3-Bench-robot}} & \multicolumn{6}{c}{\textbf{M3-Bench-web}} & \multirow{2}{*}{\begin{tabular}{@{}c@{}}\textbf{HippoVlog}\end{tabular}} \\
        \cmidrule(lr){2-7} \cmidrule(lr){8-13}
        & ME & MH & CM & PU & GK & \textbf{All} & ME & MH & CM & PU & GK & \textbf{All} & \\
        \midrule
        
        \multicolumn{14}{c}{\textit{Base Models}} \\
        \midrule
        Qwen3-VL-flash & 9.3 & 9.2 & 8.4 & 10.7 & 7.7 & \cb{9.0} & 11.9 & 10.5 & 13.4 & 14.0 & 20.9 & \cb{14.9} & \cb{24.2} \\
        Gemini-3-Pro & 15.5 & 19.8 & 14.1 & 27.1 & 7.2 & \cb{15.3} & 18.0 & 17.9 & 23.8 & 23.1 & 28.7 & \cb{23.2} & \cb{38.0} \\
        GPT-5.2 & 18.3 & 16.2 & 18.7 & 29.5 & 15.7 & \cb{18.0} & 21.3 & 21.9 & 30.9 & 27.1 & 39.6 & \cb{28.7} & \cb{38.8} \\
        \midrule
        
        \multicolumn{14}{c}{\textit{Online Video Understanding Methods}} \\
        \midrule
        MovieChat & 13.3 & 9.8 & 12.2 & 15.7 & 7.0 & \cb{11.2} & 12.2 & 6.6 & 12.5 & 17.4 & 11.1 & \cb{12.6} & \cb{32.1} \\
        MA-LLM & 25.6 & 23.4 & 22.7 & 39.1 & 14.4 & \cb{24.4} & 26.8 & 10.5 & 22.4 & 39.3 & 15.8 & \cb{24.3} & \cb{40.4} \\
        Flash-VStream & 21.6 & 19.4 & 19.3 & 24.3 & 14.1 & \cb{19.4} & 24.5 & 10.3 & 24.6 & 32.5 & 20.2 & \cb{23.6} & \cb{38.2} \\
        \midrule
        
        \multicolumn{14}{c}{\textit{Video Memory-based Methods}} \\
        \midrule
        EgoRAG & 22.6 & 27.5 & 21.0 & 37.8 &  9.2 & \cb{21.9} & 23.1 & 15.2 & 34.6 & 31.1 & 48.3 & \cb{32.8} & \cb{57.5}\\
        HippoMM & 41.2 & 40.3 & 41.5 & 56.5 & 33.1 & \cb{40.4} & 38.8 & 30.8 & 21.3 & 47.3 & 48.6 & \cb{40.7} & \cb{71.9}\\
        M3-Agent & 30.4 & 25.0 & 27.1 & 60.4 & 31.3 & \cb{41.4} & 45.9 & 28.4 & 44.3 & 59.3 & 53.9 & \cb{48.9} & \cb{65.5} \\
        WorldMM & 34.4 & 55.4 & 36.6 & 43.2 & 25.5 & \cb{34.5} & 59.2 & 42.1 & 51.7 & 69.1 & 58.4 & \cb{58.1} & \cb{71.5} \\
        \midrule

        \textbf{CAM (GPT)} & \textbf{65.5} & \textbf{80.8} & \textbf{64.8} & \textbf{82.3} & \textbf{35.3} & \cb{\textbf{64.5}} & \textbf{64.6} & \textbf{54.7} & \textbf{52.1} & \textbf{75.6} & \textbf{75.5} & \cb{\textbf{69.3}} & \cb{\textbf{76.4}} \\
        \textbf{CAM (Qwen)} & 51.2 & 63.5 & 51.5 & 70.3 & 33.4 & \cb{51.3} & 57.2 & 44.4 & 51.0 & 58.3 & 41.8 & \cb{55.6} & \cb{73.8} \\
        \bottomrule
    \end{tabular}%
    }
    \label{tab:main}
\end{table*}

Table~\ref{tab:main} reports the performance of CAM against a wide range of base models and state-of-the-art long-video understanding methods on three benchmarks: M3-Bench-robot, M3-Bench-web, and HippoVlog. Overall, CAM consistently achieves the best results across reasoning categories and datasets, demonstrating the effectiveness of hierarchical, graph-structured video memory for long-horizon reasoning.

M3-Bench-robot is the most challenging benchmark evaluated, featuring a high density of complex multi-hop questions and tasks that require deep character understanding. On tasks requiring such extended long-term dependencies, purely streaming models like MovieChat, MA-LLM, and Flash-VStream trail significantly behind memory-based methods. Accordingly, CAM obtains the largest gains over baselines on this dataset. CAM (GPT) achieves an overall accuracy of 64.5\%, outperforming the strongest prior memory-based baseline, M3-Agent, by 23.1 percentage points. Notably, even with the smaller Qwen backbone, CAM achieves a highly competitive accuracy of 51.3\%, still surpassing all baselines. Although the Qwen variant does not match the peak performance of the GPT model, it offers an efficient alternative, operating at substantially lower inference cost compared with GPT. The improvements of CAM (GPT) are especially pronounced in reasoning-intensive categories such as multi-hop reasoning (MH) and person understanding (PU). While stronger memory agents such as M3-Agent and WorldMM attempt to address these areas, their reliance on premature semantic abstraction or disjointed memory retrieval restricts their ability to support complex relational and multi-hop reasoning. In fact, WorldMM struggles noticeably on M3-Bench-robot, suggesting its segregated memory design is poorly suited to benchmarks requiring strict character consistency and identity tracking.

\begin{wrapfigure}{l}{0.4\textwidth}
    \centering
    \begin{subfigure}{\linewidth} 
        \centering
        \includegraphics[width=0.9\linewidth]{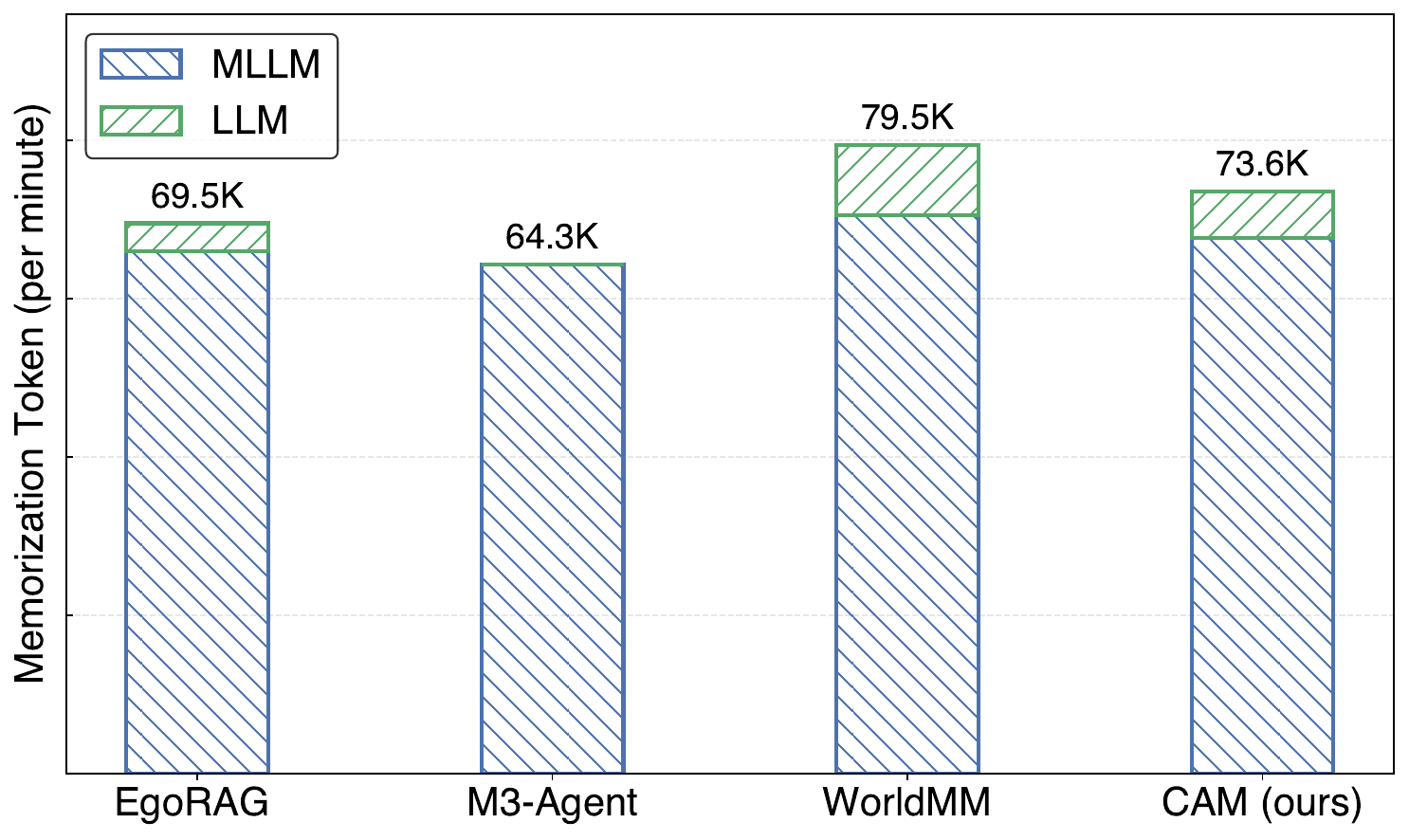}
        \caption{Memorization token consumption}
        \label{fig:token_mem}
    \end{subfigure}
    
    \vspace{1em} 
    
    \begin{subfigure}{\linewidth}
        \centering
        \includegraphics[width=0.9\linewidth]{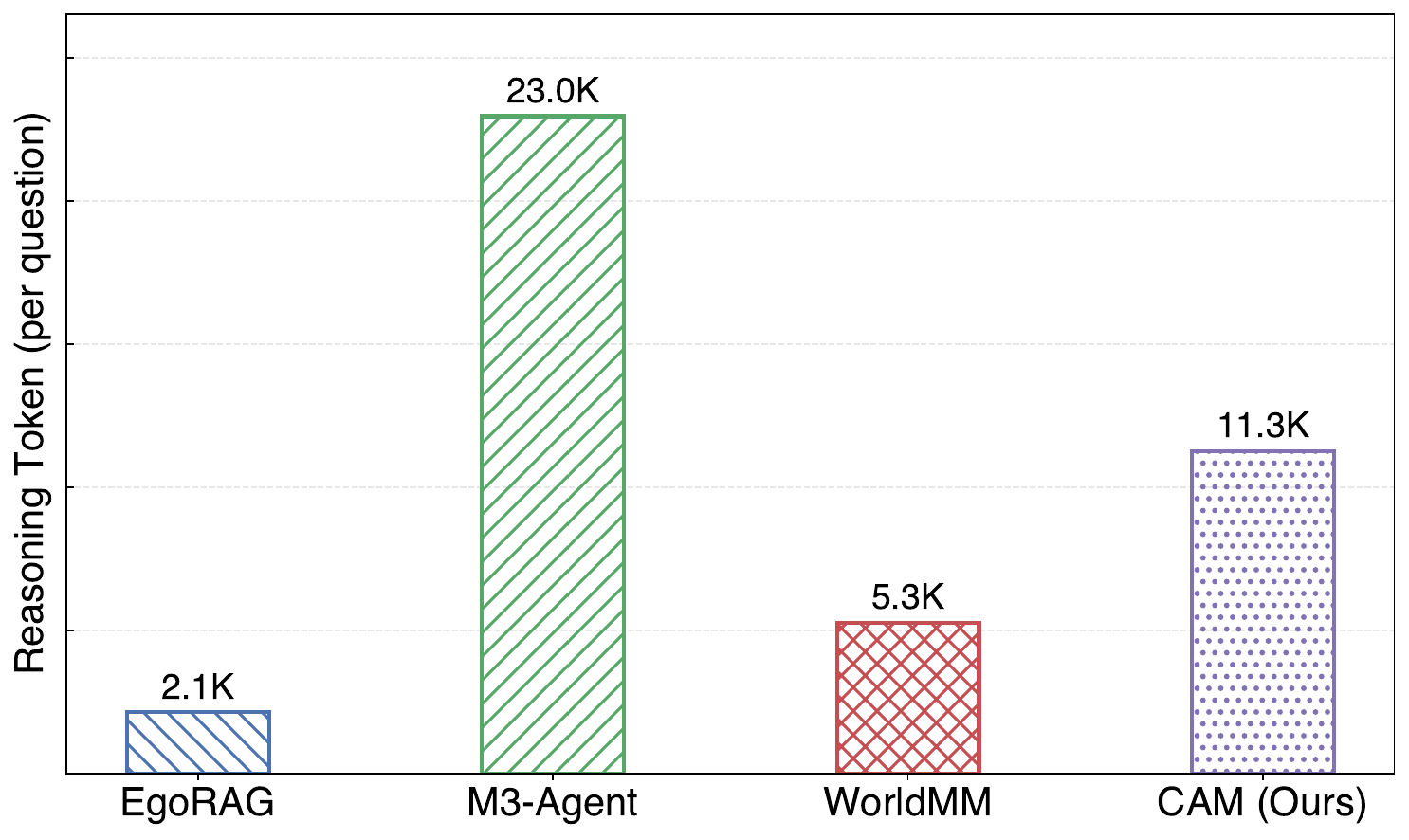}
        \caption{Reasoning token consumption}
        \label{fig:token_reason}
    \end{subfigure}
    
    \vspace{0.5em}
    \caption{Token consumption comparison. (a) Memorization tokens per minute. (b) Reasoning tokens per question.}
    \label{fig:token}
\end{wrapfigure}

On M3-Bench-web, which contains more diverse and unconstrained YouTube videos, CAM again establishes a clear lead with an overall score of 69.3\%, surpassing the strongest baseline, WorldMM, by 11.2 percentage points. Specifically, CAM shows strong improvements in person understanding (PU) and general knowledge extraction (GK), indicating its ability to integrate dispersed factual cues and identity-related information across long temporal spans. Although the overall performance gap is narrower than in the egocentric setting, CAM's consistent advantage shows that its structured memory remains robust under higher visual and semantic variability.

Furthermore, on HippoVlog, which tests continuous audiovisual understanding, CAM achieves a state-of-the-art accuracy of 76.4\%. While baselines like M3-Agent and HippoMM tend to overfit to their native benchmarks, and flat-memory methods (EgoRAG, HippoMM) suffer from lossy text compression, CAM generalizes effectively. It outperforms domain-specific models like HippoMM (71.9\%) and WorldMM (71.5\%) without any format tailoring, suggesting that its temporally grounded graph robustly captures complex vlog narratives. Detailed reasoning statistics are in Appendix~\ref{sec:statistics}.

Efficiency evaluations show that CAM maintains a favorable accuracy-cost trade-off. It incurs a moderate memorization cost of 73.6K tokens per minute, avoiding the excessive summarization overhead seen in WorldMM. At the reasoning stage, CAM utilizes 11.3K tokens per question; its structure-aware retrieval and directed tool navigation avoid the repetitive search loops that drive M3-Agent’s cost to 23.0K tokens. While CAM uses more reasoning tokens than frame-restricted baselines to preserve visual continuity, this investment is justified by its consistent accuracy gains.

\subsection{Ablation Study}

\begin{table}[!t]
\caption{Ablation study for the designs of CAM. Structured graph memory and adaptive querying tools both contribute critically to overall reasoning performance.}
\centering
\definecolor{myblue}{RGB}{232, 242, 252}
\newcommand{\cb}[1]{{\setlength{\fboxsep}{2pt}\colorbox{myblue}{\makebox[2em]{#1}}}}
\setlength{\tabcolsep}{5pt}
\resizebox{0.7\linewidth}{!}{
\begin{tabular}{lcccccc}
\toprule
 & \multicolumn{6}{c}{\textbf{M3-Bench-robot}} \\
\cmidrule(lr){2-7}
Method & ME & MH & CM & PU & GK & All \\
\midrule
\multicolumn{7}{l}{\textbf{Ablation on memory construction}} \\
Fixed-interval summarization & 62.4 & 75.8 & 61.5 & 74.2 & 32.8 & \cb{60.3} \\
CAM w/o graph      & 43.4 & 56.8 & 47.9 & 57.9 & 27.3 & \cb{44.8} \\
CAM w/o high-level & 54.6 & 66.9 & 51.5 & 79.9 & 29.1 & \cb{55.7} \\
\midrule
\multicolumn{7}{l}{\textbf{Ablation on question answering}} \\
similarity search only    & 56.6 & 73.1 & 57.2 & 80.2 & 34.3 & \cb{56.5} \\
general search + video rewatch & 59.5 & 75.3 & 58.8 & 82.3 & 29.2 & \cb{57.8} \\
decreasing top-k to 30 & 63.4 & 75.1 & 62.1 & 80.4 & 34.7 & \cb{62.9} \\
\midrule
\textbf{CAM-full} & \textbf{65.5} & \textbf{80.8} & \textbf{64.8} & \textbf{82.3} & \textbf{35.3} & \cb{\textbf{64.5}} \\
\bottomrule
\end{tabular}
}
\vspace{-2mm}
\label{tab:ablation}
\end{table}
We conduct ablations to examine CAM's memory construction and question-answering designs. 
For memory construction, we compare CAM with three variants: 
(1) \textbf{CAM w/o graph}, which removes the heterogeneous graph and reduces retrieval to unstructured semantic matching; 
(2) \textbf{CAM w/o high-level}, which keeps only low-level episodic edges and disables entity- and relation-level summaries; and 
(3) \textbf{fixed-interval abstraction}, which generates high-level summaries every 10 minutes instead of using evidence-driven subgraph maturity.
For question answering, we evaluate three constrained reasoning settings: 
(1) similarity search only, 
(2) general search with video rewatch, and 
(3) a reduced search budget that decreases top-$k$ from 50 to 30.

As shown in Table~\ref{tab:ablation}, removing the graph structure causes the largest drop, reducing the overall score from 64.5\% to 44.8\%. 
The degradation is especially severe on Multi-Hop (80.8\% to 56.8\%) and Multi-Evidence reasoning (65.5\% to 43.4\%), showing that unstructured memory and similarity matching cannot effectively capture the spatio-temporal dependencies in long videos.
Removing high-level abstraction also hurts performance, dropping the overall score to 55.7\%, with notable declines on Cross-Modal and Multi-Hop questions. 
This indicates that episodic observations alone are insufficient, and entity- or relation-level summaries are necessary for connecting scattered evidence into coherent semantics.
Moreover, fixed-interval abstraction reaches only 60.3\%, confirming that summarizing at arbitrary temporal boundaries can introduce premature or unreliable abstractions, whereas CAM's evidence-driven abstraction yields more mature and consistent high-level memory.

For question answering, restricting the agent to similarity search yields 56.5\%, and adding video rewatch only slightly increases it to 57.8\% while significantly raising token costs. 
These results confirm that naive retrieval cannot replace CAM's specialized graph-aware tools, such as object search and graph traversal. 
Finally, reducing the search budget to top-$k=30$ causes a minor drop to 62.9\%, showing robustness under tighter retrieval constraints while benefiting from wider coverage for optimal reasoning.

\section{Conclusion}
\label{sec:conclusion}
In this paper, we study reliable memory construction for complex question answering over long-form videos. We propose CAM, a structured video memory framework that organizes clip-level observations into a temporally grounded heterogeneous graph. 
By continuously abstracting mature subgraphs into high-level semantics and adaptively invoking graph-based and multimodal retrieval tools, CAM bridges the gap between low-level evidence and high-level reasoning, achieving strong performance on complex multi-hop entity-centric video QA benchmarks.

\bibliographystyle{unsrtnat}
\bibliography{HVM}

\clearpage

\appendix 

\begin{center}
    \Large \textbf{Appendix Contents}
\end{center}
\vspace{-0.5em}
\vspace{0.5em}

\startcontents[sections]
\printcontents[sections]{l}{1}{\setcounter{tocdepth}{2}}

\vspace{0.5em}
\noindent\rule{\textwidth}{0.4pt} 
\vspace{2em}

\section{Experimental Setup and Implementation Details}
\label{sec:implementation}

\subsection{Datasets and Metrics}
We evaluate our method on three long-video understanding benchmarks: \textbf{M3-Bench-robot}, \textbf{M3-Bench-web}, and \textbf{HippoVlog}. M3-Bench-robot contains 100 long-form videos and 1,276 questions. The videos have an average duration of 34 minutes, are captured from robotic viewpoints, and cover five reasoning categories: multi-evidence reasoning (ME), multi-hop reasoning (MH), cross-modal reasoning (CM), person understanding (PU), and general knowledge extraction (GK). Performance on M3-Bench-robot is evaluated using accuracy, with predictions judged by GPT-5.2. M3-Bench-web follows the same reasoning categories and evaluation protocol as M3-Bench-robot, but is composed of YouTube videos with an average length of 27 minutes. Finally, HippoVlog consists of 25 videos with an average duration of 27.2 minutes and is evaluated using 1,000 multiple-choice questions across four distinct categories: cross-modal (V+A), auditory (A), visual (V), and semantic (S).

\stitle{LLM judge protocol and validation}
For M3-Bench, we reuse the official M3-Agent evaluator prompt~\cite{m3agent}, which determines whether the reference answer can be inferred from a prediction and returns only \texttt{"Yes"} or \texttt{"No"}. Predictions follow the benchmark's concise one-sentence answer format. Reported scores use GPT-5.2 as the judge. GPT-5.2 and the GPT-4o cross-judge receive the same prompt in separate post-generation calls, and the full prompt is provided in Appendix~\ref{sec:answer_verification}. The answering agent never receives the evaluator prompt, reference answer, or evaluator output during reasoning.

To validate the judge, we independently evaluate all 1,276 M3-Bench-robot predictions with GPT-4o and human annotations:
\begin{itemize}[leftmargin=*, topsep=2pt, itemsep=0pt, parsep=0pt]
    \item \textbf{GPT-5.2 vs. GPT-4o:} 96.7\% agreement.
    \item \textbf{GPT-5.2 vs. human:} 97.7\% agreement.
\end{itemize}
The resulting accuracies under GPT-5.2, GPT-4o, and human evaluation are 64.5\%, 67.2\%, and 65.0\%, respectively. The human-evaluated accuracy is only 0.5 percentage points above the reported GPT-5.2-judged result, while GPT-4o yields a higher score. These results provide no evidence that using GPT-5.2 to judge answers generated by the same model family inflates the reported accuracy. M3-Agent also reports 96\% agreement between GPT-4o and human majority vote in Section~3.3 and Table~18 of its paper~\cite{m3agent}. HippoVlog uses exact matching for its multiple-choice answers and therefore requires no LLM judge.

\subsection{Baselines}
We compare CAM against a comprehensive set of baselines spanning base video LLMs, long video understanding models, RAG systems, and memory-based models. Base video LLMs include GPT-5.2, Gemini-3-Pro, and Qwen3-VL-flash, while long video understanding models include MovieChat, MA-LLM, Flash-VStream. Finally, we compare with memory-based frameworks for long video reasoning, including EgoRAG, HippoMM, M3-Agent, and WorldMM:

\begin{itemize}[leftmargin=*, topsep=0pt, itemsep=0pt, parsep=0pt]
    \item \textbf{EgoRAG:} EgoRAG \cite{egolife} segments videos into short clips and generates hierarchical text captions based on rigid temporal boundaries. To answer questions, it searches these captions using text similarity. While EgoRAG is designed for lifelong egocentric understanding and continuous memory accumulation, its memory structure primarily serves basic retrieval and summarization rather than complex, explicit entity-relation reasoning.
    
    \item \textbf{HippoMM:} HippoMM utilizes a biologically inspired episodic-semantic memory structure to improve the long-term retention of video content. However, because its memory organization is largely sequential, the model is limited in its ability to effectively handle multi-hop and relational reasoning tasks. It is treated as a special case and exclusively uses Qwen models instead of GPT. This is because HippoMM involves extensive preprocessing steps, including dynamic segmentation and consolidation. Using GPT combined with the required GPU processing for these steps would be highly computationally expensive. Furthermore, it is important that the processing time remains shorter than the length of the video itself, which GPT's overhead would prevent. Consequently, its original experimental setting is maintained.
    
    \item \textbf{M3-Agent:} This framework explicitly models entities and organizes extracted facts around them to construct an entity-centric memory. It works by storing clip-level textual summaries within an external memory. The original M3-Agent framework was trained using reinforcement learning with a relatively small base model (Qwen2.5-omni-7B). Because this small base model lacks both generality and the capability of handling complex visual tasks across other benchmarks, we evaluate M3-Agent using GPT-5.2 to ensure a fair comparison with the other methods. Its main drawback is that its retrieval mechanism underutilizes graph topology; it relies on flat textual similarity matching, which struggles to navigate complex relational chains.
    
    \item \textbf{WorldMM:} WorldMM \cite{worldmm} attempts to capture both temporal events and higher-level semantics by maintaining three separate memory modules: episodic, semantic, and visual. Despite this comprehensive approach, it physically segregates its knowledge into decoupled memory banks. Because the visual features are isolated from the textual knowledge graph, the agent is forced to conduct independent, fragmented retrieval streams instead of achieving a unified multimodal understanding.
\end{itemize}

\subsection{Implementation Details}
CAM and all baseline methods (except the three base models and HippoMM) adopt GPT-5.2 as the backbone for both text-only and multimodal processing to ensure a fair comparison. To evaluate the effect of different LLMs on the video memory frameworks, CAM additionally uses Qwen3-VL-flash (MLLM) and Qwen3.5-flash (LLM). HippoMM is treated as a special case and exclusively uses the Qwen models. Because HippoMM involves extensive preprocessing steps, including dynamic segmentation and consolidation, using the GPT model would make the process prohibitively slow, so its original experimental setting is maintained. For video input, EgoRAG, M3-Agent, WorldMM, and CAM all use 30-second fixed-interval clips, whereas HippoMM relies on dynamic video segmentation with clips ranging from 5 to 10 seconds long. Audio streams are transcribed and incorporated into the video representation as subtitles, enabling unified multimodal reasoning, and videos are uniformly sampled at 1 fps. During dialogue retrieval, the sentences immediately preceding and following the matched conversation are also included in the context window, as answers often rely on surrounding conversational context rather than the retrieved utterance alone.

\section{Additional Method Details}
\label{sec:appendix_method}

\subsection{Natural-Language Entity Tracking}

While prior methods such as M3-Agent~\cite{m3agent} use InsightFace embeddings for face association, CAM extracts and passes appearance descriptors ($A_{i-1}$) as natural language. Visual-embedding trackers rely primarily on local appearance similarity and can become unreliable under occlusion, illumination changes, extreme poses, or blurred faces. CAM's streaming MLLM can instead combine clothing, surrounding people and objects, neighboring frames, and previously processed context. When the visual evidence in one clip is ambiguous, later clips can also resolve an earlier provisional identifier through equivalence updates.

\stitle{Association Quality} We manually annotated 13 videos spanning 6.98 hours and 846 clips, establishing 47 distinct characters as ground-truth identities. We ran the released M3-Agent implementation on the same videos using its provided InsightFace parameters without retuning, then aligned its predicted identity clusters with the ground truth. Table~\ref{tab:entity_association} summarizes the results.

\begin{table}[H]
    \centering
    \caption{Character-association errors on the manually annotated subset. ``Merge/miss'' combines false merges and missed identities. Added latency is measured per 30-second clip on an RTX 4090.}
    \label{tab:entity_association}
    \setlength{\tabcolsep}{7pt}
    \begin{tabular}{@{} l c c c c @{}}
        \toprule
        \textbf{Association method} & \textbf{Total errors} & \textbf{Splits} & \textbf{Merge/miss} & \textbf{Added latency} \\
        \midrule
        CAM natural-language association & \textbf{14} & 14 & \textbf{0} & \textbf{0 s} \\
        InsightFace & 18 & 1 & 17 & 8.5 s \\
        \bottomrule
    \end{tabular}
\end{table}

Relative to the ground truth, CAM splits 14 of the 47 identities once each, creating 14 excess nodes, but produces no false merges or missed identities. Five of the 13 videos are fully correct, and the resulting fragments contain only 5.2\% of all graph edges. InsightFace produces 17 false-merge or missed-identity errors and one split, for a total of 18 association errors. Thus, CAM yields fewer errors overall, and its errors are limited to splits. This distinction matters for graph memory: false merges contaminate an identity with another character's actions, dialogue, and relationships, while missed identities discard evidence entirely. Splits fragment and delay evidence accumulation but do not transfer evidence across identities.

\stitle{Computational Cost and Threshold Sensitivity} On an RTX 4090, InsightFace embedding and clustering add 8.5 seconds per 30-second clip, corresponding to a 39.2\% overhead relative to CAM's 21.7-second construction time. On CPU, the same association stage takes more than one hour for a 30-minute video. Its similarity threshold also controls the split--merge trade-off. Because M3-Agent's provided setting is validated only on its native benchmark, its transferability to other visual conditions is unclear. In contrast, CAM generates appearance descriptors during the existing MLLM extraction step, requiring no additional GPU model, inference pass, or similarity-threshold tuning.

\subsection{Evidence-Driven Abstraction}
\label{subsec:appendix_abstract}

Semantic graphs from long videos differ sharply from text-derived graphs. In video, a few persistent entities—especially characters—recur across many clips, producing high‑degree nodes, dense temporal links, and strict ordering constraints. Unlike text, where facts are discrete, video evidence is incremental, noisy, and unevenly distributed. This uneven accumulation means abstraction cannot simply follow fixed time intervals. Some entities stabilize early, while others remain sparse until much later. Premature abstraction risks inconsistency, while delayed abstraction wastes evidence already sufficient for reasoning. Both mismatches undermine memory coherence and downstream analysis.

Consequently, fixed-interval abstraction misaligns with evidential support, yielding premature or delayed summaries that undermine memory consistency and downstream reasoning.
To address this imbalance, CAM employs \emph{evidence-driven abstraction}, where abstractions are triggered by the accumulation of clip-level relations in the memory graph rather than periodic time steps.

Concretely, CAM applies two complementary abstractions: \emph{character-level} abstraction operates on individual nodes, triggering summaries only after sufficient incident evidence accumulates to support stable inference of long-term behaviors, whereas \emph{relationship-level} abstraction operates on node pairs, triggering only after sufficient interaction evidence accrues to capture persistent social dynamics beyond isolated characters. 
Both abstractions are conditioned on graph structure and relation density via predefined evidence thresholds, ensuring summaries are grounded in adequate observations.

\stitle{Character-Level Abstraction} For each character node $v \in V_{\text{char}}$, we maintain the set of incident edges that aggregate all episodic evidence involving that character, including actions, interactions with other entities, and associated contextual states. At processing step $t$, this set is
\[
\mathcal{E}_t(v) = \{ e \in E_t \mid e.\text{source} = v \ \text{or} \ e.\text{target} = v \}.
\]
This set grows incrementally as new clips are processed. To support repeated refinement without retriggering after every subsequent edge, we record $b_{\text{char}}(v)$, the number of incident edges present immediately after the most recent character-level abstraction, initialized to zero. Character-level abstraction is triggered only after at least $\tau_{\text{char}}$ new incident edges have accumulated:
\[
\text{Trigger}^{(t)}_{\text{char}}(v)
= \mathbb{I}\big(|\mathcal{E}_t(v)|-b_{\text{char}}(v) \ge \tau_{\text{char}}\big),
\]
where $\mathbb{I}(\cdot)$ is the indicator function.

This thresholding reflects the intuition that stable character attributes can only be inferred from \emph{sufficiently diverse observations}. 
When the trigger fires, the newly accumulated evidence, including its temporal and contextual annotations, is summarized together with the existing high-level characterization, if one is available. The LLM output updates the character-level abstract node or attribute edge linked to $v$, and the counter is advanced to $b_{\text{char}}(v)\leftarrow|\mathcal{E}_t(v)|$. All low-level episodic edges are retained for evidence preservation and future refinement.

\stitle{Relationship-Level Abstraction} Beyond individual traits, long-term video understanding requires modeling persistent inter-character relations (e.g., cooperation or rivalry) that emerge only through repeated interactions over time rather than within isolated clips.

For a character pair $(u,v)\in V_{\text{char}}\times V_{\text{char}}$, we define the relationship evidence available at processing step $t$ as
\[
\mathcal{E}_t(u,v)=\mathcal{E}_{t,\text{direct}}(u,v)\cup\mathcal{E}_{t,\text{indirect}}(u,v),
\]
where $\mathcal{E}_{t,\text{direct}}(u,v)$ contains edges corresponding to explicit interactions between $u$ and $v$, and $\mathcal{E}_{t,\text{indirect}}(u,v)$ captures co-involvement with the same object, location, or third entity within a clip or short temporal window, encoding implicit relational signals. We similarly maintain $b_{\text{rel}}(u,v)$, the relationship-evidence count recorded after the most recent abstraction, initialized to zero. Relationship abstraction is triggered after $\tau_{\text{rel}}$ new interaction edges accumulate:
\[
\text{Trigger}^{(t)}_{\text{rel}}(u,v)
=\mathbb{I}\big(|\mathcal{E}_t(u,v)|-b_{\text{rel}}(u,v)\ge\tau_{\text{rel}}\big).
\]

When the threshold $\tau_{\text{rel}}$ is reached, the newly accumulated interaction evidence is summarized together with the existing relational abstraction, if one is available. The resulting description updates the abstract relational edge between $u$ and $v$, and the counter is advanced to $b_{\text{rel}}(u,v)\leftarrow|\mathcal{E}_t(u,v)|$.

As with character-level abstraction, low-level relational edges are preserved to allow subsequent refinement as new interactions are observed. By conditioning each update on a fresh batch of accumulated relational evidence, CAM ensures that relationship summaries reflect consistent behavioral patterns rather than transient or coincidental interactions.

\subsection{Agent Toolbox}
\label{subsec:tool}

To empower the agent to actively navigate the heterogeneous memory graph and ground its reasoning in raw multimodal signals, we define a comprehensive suite of seven specialized tools. Each tool is designed to overcome specific bottlenecks in long-form video understanding, ranging from structured knowledge retrieval to raw sensory verification.

\begin{itemize}[leftmargin=*, topsep=0pt, itemsep=0pt, partopsep=0pt]
    \item \textbf{General Search (\texttt{general\_search}):} This tool serves as the agent's foundational retrieval mechanism. To address queries that require both global entity relations and fine-grained visual or conversational details, this search operation is mathematically decomposed into three distinct stages:
    
    \vspace{1mm}
    \noindent\textit{1) Weighted Query Decomposition.} 
    During tool invocation, the Planner Node directly parses the natural-language query $q$ into a semantic triple $\mathcal{T}_q = \langle s_q, r_q, t_q \rangle$. To capture the relative importance of each component, the Planner dynamically assigns a normalized weight vector $\mathbf{W}_q = \langle w_s, w_r, w_t \rangle$ (where $w_s + w_r + w_t = 1$). Simultaneously, it extracts auxiliary context $I_q$ and determines a retrieval budget $\mathbf{K}_q = \langle k_{\text{high}}, k_{\text{low}}, k_{\text{conv}} \rangle$ to allocate how many nodes to retrieve across the three memory channels. These parameters collectively form the tool's execution arguments:
    \begin{equation}
        \text{args}_t = \big(\mathcal{T}_q,\, \mathbf{W}_q,\, \mathbf{K}_q,\, I_q \big)
    \end{equation}

    \vspace{1mm}
    \noindent\textit{2) Triple Similarity Computation.} 
    To evaluate relevance within the heterogeneous graph $\mathcal{G}$, we define a similarity function in a shared latent space mapped by a dense embedding function $E$. The base cosine similarity between elements is $sim(x, y) = \frac{E(x)^\top E(y)}{\lVert E(x) \rVert_2 \lVert E(y) \rVert_2}$. When comparing the query triple $\mathcal{T}_q = \langle s_q, r_q, t_q \rangle$ to a candidate triple $\mathcal{T}_c = \langle s_c, r_c, t_c \rangle \in \mathcal{G}$, we first account for reversed semantic roles using a weighted directional entity alignment score for the nodes:
    $$
    \text{Align}\big((s_q, t_q), (x, y)\big) = w_s \cdot sim(s_q, x) + w_t \cdot sim(t_q, y)
    $$
    The total structural similarity, $S_{\text{struct}}$, integrates this bidirectional node alignment with the direct semantic matching of the relation, ensuring overall robustness against role reversals:
    $$
    S_{\text{struct}}(\mathcal{T}_q, \mathcal{T}_c) = \max\!\Big( \text{Align}\big((s_q,t_q),(s_c,t_c)\big),\; \text{Align}\big((s_q,t_q),(t_c,s_c)\big) \Big) + w_r \cdot sim(r_q, r_c)
    $$

    \vspace{1mm}
    \noindent\textit{3) Multi-Channel Retrieval.} 
    Retrieval is executed independently across partitioned memory channels to reflect their distinct semantic roles. High-level abstractions are retrieved primarily via structural similarity to capture long-term, aggregated knowledge. Low-level episodic relations require fine-grained spatio-temporal grounding, thus incorporating the extracted context $I_q$ alongside structural similarity ($S_{\text{struct}} + \lambda \cdot sim(I_q, C)$). Conversely, conversational content bypasses structured triples entirely, relying on direct natural-language matching against the original query ($\mathrm{arg\,max}_i \, sim(q, d_i)$). To preserve discursive coherence, retrieved utterances are expanded to include their immediate chronological neighbors ($d_{i-1}, d_{i+1}$). Finally, all retrieved evidence is merged into a unified set $\mathcal{R} = \mathcal{R}_{\text{high}} \cup \mathcal{R}_{\text{low}} \cup \mathcal{R}_{\text{conv}}$, enabling the system to flexibly integrate abstract, episodic, and dialogue contexts for downstream reasoning.

    \vspace{1mm}
    \item \textbf{Temporal Context Search (\texttt{search\_temporal\_context}):} For queries requiring strict temporal localization, this tool extends the \texttt{general\_search} operation by introducing two bounding parameters: $t_{\text{begin}}$ and $t_{\text{end}}$. By constraining the search space to a specific temporal window, the agent effectively prunes irrelevant historical or future events, focusing the similarity computation exclusively on a localized subgraph.

    \item \textbf{Search Object (\texttt{search\_object}):} Visual perception models frequently introduce vocabulary discrepancies during initial graph construction (e.g., misclassifying a ``marker'' as a ``pen''). To mitigate this semantic mismatch, this tool bypasses rigid lexical matching in favor of robust, embedding-based entity alignment, ensuring accurate object identification despite variations in natural language taxonomy.

    \item \textbf{Frequency Statistics (\texttt{get\_frequency\_stats}):} A quantitative tool that aggregates occurrences across the timeline, returning the exact number of times a specific action, object, or event appears in the structured memory.

    \item \textbf{Graph Traversal (\texttt{graph\_traversal}):} Enables explicit topological traversal of the memory structure for multi-hop deduction. Let $e_a \in \mathcal{E}$ denote an anchor entity currently held in the agent's intermediate memory $\mathcal{I}_t$. Given a target relation query $r_q$, this tool retrieves the semantically filtered neighborhood: $\mathcal{N}(e_a, r_q) = \{ e_b \in \mathcal{E} \mid \exists \langle e_a, r, e_b \rangle \in \mathcal{G} \text{ s.t. } sim(r_q, r) \geq \tau \}$, where $\tau$ is a similarity threshold. By sequentially applying this operation, the agent actively constructs a deductive path $\mathcal{P} = (e_0, r_1, e_1, \dots, r_k, e_k)$ across heterogeneous edges (e.g., character relationships or causal links), bridging disparate temporal video segments strictly through graph topology rather than global search.

    \item \textbf{Video Rewatch (\texttt{watch\_video\_clip}):} The initial partitioning of video into textual triples via MLLMs inevitably compresses subtle visual nuances, such as background context and continuous motion dynamics. Serving as a critical multimodal fallback mechanism, this tool accepts a specific clip identifier as input, enabling the agent to bypass the structural memory and directly retrieve the raw visual frames for re-evaluation. This ensures the reasoning process remains firmly grounded in empirical visual evidence when the generated graph lacks the fine-grained details necessary to resolve semantic ambiguities.

    \item \textbf{Audio Listening (\texttt{listen\_to\_audio}):} Similarly, non-speech acoustic cues (e.g., environmental sound effects or background noise) are frequently omitted during graph construction, as transcription models predominantly focus on character dialogue. This tool allows the agent to bypass the text modality and directly sample the raw audio track within targeted clips, providing essential auditory grounding when the transcribed contextual information is insufficient.
\end{itemize}

\section{Additional Related Work}
\label{sec:appendix_related_work}

\stitle{Long Video Understanding}
Long video understanding remains challenging due to limited context windows, evolving semantics, and the need to preserve both fine-grained details and long-range temporal dependencies~\cite{mesh}, motivating memory-centric designs that compress or store video information for long-horizon reasoning, such as compact textual summaries, fixed-size streaming memories, or hierarchical memory across abstraction levels with bidirectional or event-level updates~\cite{moviechat,ma_lmm,qian2024streaming,wu2024bidirectional,balazevic2024memory,rewind,flashvstream,videotree,timechat}. While effective, these methods often rely on fixed memory budgets or heuristic updates, leading to information loss when critical details appear late in the video, whereas retrieval-augmented approaches dynamically fetch relevant segments or auxiliary signals at query time to improve scalability~\cite{luo2024videorag,gia2025vrag,goldfish}, but remain highly dependent on retrieval accuracy and typically treat evidence as unstructured context, limiting temporal and relational modeling. More recently, agentic and multi-agent frameworks decompose long-video reasoning into iterative or collaborative subtasks, with VideoAgent~\cite{videoagent}, VideoMultiAgents~\cite{videomultiagents}, and LongVideoAgent~\cite{longvideoagent} enabling targeted or parallel analysis through tool use and shared memory, yet overall existing methods still trade off between memory compression, retrieval efficiency, and reasoning capacity, motivating unified frameworks that integrate memory, retrieval, and structured reasoning for robust long-horizon understanding.

\stitle{Graph-based Video Search}
Graph structures have been widely explored in text-based retrieval-augmented generation and memory systems to model relational knowledge and support multi-hop reasoning~\cite{edge2024graphrag,zhou2024atom,anokhin2025arigraph,rasmussen2025zep, recmem}. In the video domain, structured representations have increasingly been adopted to address the limitations of fixed context windows and sequential processing, with early spatio-temporal graphs explicitly modeling object interactions and temporal dynamics for activity and event understanding~\cite{chu2025entitygraph, videomultiagents}. More recent approaches leverage graph-based memories and prompting to bridge high-level semantics and low-level video content, such as Glance-Focus~\cite{bai2023glance}, while explicit entity tracking further supports long-form reasoning through evolving relational graphs, as in GraphVideoAgent, which achieves strong performance on long-horizon benchmarks including EgoSchema and NExT-QA~\cite{chu2025entitygraph}. Structured retrieval-augmented frameworks combine graphs with selective retrieval to scale to extreme video lengths, including VideoRAG’s graph-grounded semantic indexing~\cite{VideoRAG}, RAVU’s compositional reasoning over spatio-temporal entity graphs~\cite{malik2025ravu}, and Vgent’s graph-based retrieval–reasoning–augmented generation with intermediate verification~\cite{shen2025vgent}. Despite these advances, most methods rely on relatively flat relational structures, treat multimodal signals as external augmentations rather than unified graph components, and lack explicit modeling of hierarchical event abstractions or causal dependencies across clips, motivating richer and more integrated graph representations for robust long-video understanding.

\stitle{Memory-Based Video LLMs}
To support long-video reasoning, memory-based methods construct structured knowledge over video streams. EgoLife \cite{egolife} emphasizes lifelong egocentric understanding with continuous memory accumulation, but its memory mainly serves retrieval and summarization rather than explicit entity–relation reasoning; Ego-R1 \cite{egor1} extends this direction with vision-centric tools and iterative reasoning for long-horizon tasks, yet still lacks a structured relational representation. HippoMM \cite{hippomm} adopts a biologically inspired episodic–semantic memory to improve long-term retention, but its largely sequential organization limits multi-hop and relational reasoning, while M3-Agent \cite{m3agent} stores clip-level textual summaries in an external memory, leading to unstructured representations and premature abstraction that are difficult to revise as new evidence emerges.

In contrast, CAM models long-video memory as a heterogeneous, temporally grounded semantic graph, deferring high-level abstraction until sufficient evidence accumulates and separating episodic observations from semantic summaries. Its multi-channel retrieval and query-guided video rewatching further enable efficient access to both abstract knowledge and raw visual evidence, supporting robust reasoning when critical information is sparsely distributed over time.

\section{Additional Experiments and Analysis}
\label{sec:statistics}

\subsection{Abstraction Threshold Sensitivity}

Unlike temporal boundaries, which force abstraction regardless of content density, the thresholds $\tau_{\mathrm{char}}$ and $\tau_{\mathrm{rel}}$ are grounded in information density. In our implementation, these thresholds are empirically set to $\tau_{\mathrm{char}}=50$ and $\tau_{\mathrm{rel}}=30$. A threshold that is too low causes premature abstraction, whereas a threshold that is too high delays the emergence of high-level semantics and risks exceeding the model's context window during summarization. Table~\ref{tab:threshold_sensitivity} evaluates the sensitivity of CAM to these thresholds.

\begin{table}[H]
    \centering
    \caption{Accuracy (\%) under different abstraction settings on M3-Bench-robot and M3-Bench-web. Threshold pairs denote $\tau_{\mathrm{char}}/\tau_{\mathrm{rel}}$; the fixed-interval variant does not use evidence thresholds.}
    \label{tab:threshold_sensitivity}
    \definecolor{thresholdblue}{RGB}{232, 242, 252}
    \newcommand{\thresholdcb}[1]{{\setlength{\fboxsep}{2pt}\colorbox{thresholdblue}{\makebox[2em]{#1}}}}
    \setlength{\tabcolsep}{3pt}
    \resizebox{\linewidth}{!}{%
    \begin{tabular}{@{} l ccccc c ccccc c @{}}
        \toprule
        \multirow{2}{*}{\textbf{Setting ($\tau_{\mathrm{char}}/\tau_{\mathrm{rel}}$)}} & \multicolumn{6}{c}{\textbf{M3-Bench-robot}} & \multicolumn{6}{c}{\textbf{M3-Bench-web}} \\
        \cmidrule(lr){2-7} \cmidrule(lr){8-13}
        & ME & MH & CM & PU & GK & \textbf{All} & ME & MH & CM & PU & GK & \textbf{All} \\
        \midrule
        Fixed-interval & 62.4 & 75.8 & 61.5 & 74.2 & 32.8 & \thresholdcb{60.3} & 58.6 & 50.0 & 56.3 & 75.6 & 71.7 & \thresholdcb{66.9} \\
        $30/10$ & 63.6 & 73.6 & 63.6 & 81.9 & 33.0 & \thresholdcb{61.9} & 63.6 & \textbf{57.8} & 47.9 & 74.0 & 67.9 & \thresholdcb{67.5} \\
        $50/30$ (default) & \textbf{65.5} & \textbf{80.8} & \textbf{64.8} & 82.3 & \textbf{35.3} & \thresholdcb{\textbf{64.5}} & \textbf{64.6} & 54.7 & \textbf{52.1} & 75.6 & \textbf{75.5} & \thresholdcb{\textbf{69.3}} \\
        $100/60$ & 63.7 & 74.8 & 64.6 & \textbf{82.8} & 35.0 & \thresholdcb{63.1} & 61.6 & 51.6 & 50.0 & \textbf{76.3} & 73.6 & \thresholdcb{68.1} \\
        \bottomrule
    \end{tabular}
    }
\end{table}

\subsection{Reasoning Efficiency}

\begin{table*}[!htbp]
    \centering
    \resizebox{\textwidth}{!}{%
    \begin{tabular}{@{} l ccc ccc ccc @{}}
        \toprule
        \multirow{2}{*}{\textbf{Method}} & \multicolumn{3}{c}{\textbf{M3-Bench-robot}} & \multicolumn{3}{c}{\textbf{M3-Bench-web}} & \multicolumn{3}{c}{\textbf{HippoVlog}} \\
        \cmidrule(lr){2-4} \cmidrule(lr){5-7} \cmidrule(lr){8-10}
        & Avg rounds & correct & wrong & Avg rounds & correct & wrong & Avg rounds & correct & wrong \\
        \midrule
        M3-Agent & 3.14 & 2.55 & 3.58 & 3.11 & 2.68 & 3.52 & 2.76 & 2.46 & 3.33 \\
        WorldMM  & 2.74 & 1.85 & 3.22 & 2.21 & 2.08 & 2.41 & 1.28 & 1.2 & 1.49 \\
        CAM      & 2.2  & 1.78 & 2.92 & 2.48 & 2.36 & 2.71 & 2.13 & 2.03 & 2.41 \\
        \bottomrule
    \end{tabular}%
    }
    \caption{Average reasoning rounds required by different methods across benchmarks, separated by correct and incorrect predictions.}
    \label{tab:reasoning_rounds}
\end{table*}

Table \ref{tab:reasoning_rounds} presents a detailed breakdown of the average reasoning rounds executed by the agents, separated into correct and incorrect predictions across the three benchmarks. A consistent trend across all methods is that incorrect predictions require significantly more reasoning rounds than correct ones, illustrating the agents' tendency to exhaustively search their memory when critical evidence is missing or ambiguous. M3-Agent consistently exhibits the highest average reasoning rounds across all datasets (e.g., 3.14 on M3-Bench-robot) and struggles significantly with termination, as evidenced by its disproportionately high round counts for incorrect predictions. This aligns with its unstructured memory format, which forces the agent into repetitive, inefficient search loops when it cannot locate necessary details. In contrast, CAM demonstrates a highly efficient reasoning trajectory. By leveraging its structured, temporally grounded graph and explicit tool navigation, CAM consistently arrives at correct answers using fewer rounds than M3-Agent (e.g., requiring only 1.78 rounds versus 2.55 on M3-Bench-robot) while also avoiding endlessly repetitive loops when the required information is ultimately unavailable. Meanwhile, WorldMM shows varying round counts—dropping notably low on HippoVlog (1.28 average)—reflecting its heavy reliance on disjointed, single-pass memory bank lookups rather than sustained, multi-hop deduction.

\subsection{Tool-Use Statistics}

\begin{table*}[!htbp]
    \centering
    \small 
    \renewcommand{\arraystretch}{1.2}
    \resizebox{\textwidth}{!}{%
    \begin{tabular}{@{} l ccc @{}}
        \toprule
        \textbf{Benchmark} & \textbf{1st Most Common (\%)} & \textbf{2nd Most Common (\%)} & \textbf{3rd Most Common (\%)} \\
        \midrule
        M3-Bench-robot & \texttt{general\_search} (70.2\%) & \texttt{watch\_video\_clip} (16.8\%) & \texttt{search\_temporal\_context} (8.6\%) \\
        M3-Bench-web   & \texttt{general\_search} (78.6\%) & \texttt{search\_temporal\_context} (13.9\%) & \texttt{watch\_video\_clip} (3.7\%) \\
        HippoVlog      & \texttt{general\_search} (73.8\%) & \texttt{watch\_video\_clip} (13.3\%) & \texttt{listen\_to\_audio} (7.4\%) \\
        \bottomrule
    \end{tabular}
    }
    \caption{Distribution of the top three most frequently invoked tools by the agent during the reasoning process across different benchmarks.}
    \label{tab:tool_frequency}
\end{table*}

Table~\ref{tab:tool_frequency} details the distribution of tool calls during the agent's reasoning process across the evaluated benchmarks. The variation in tool invocation frequency directly reflects the differing question distributions and specific reasoning demands inherent to each dataset. Across all benchmarks, \texttt{general\_search} consistently ranks as the most frequently invoked tool, accounting for over 70\% of total calls. This dominance is expected, as the agent is strictly constrained to utilize \texttt{general\_search} during its initial reasoning round to establish a foundational knowledge context. Furthermore, \texttt{watch\_video\_clip} emerges as a critical secondary mechanism, ranking second for both M3-Bench-robot and HippoVlog. The prominence of this multimodal fallback tool highlights its vital role in recovering fine-grained visual details, effectively mitigating the inevitable information loss that occurs when raw video segments are initially compressed into textual representations. Finally, dataset-specific requirements dictate the remaining tool preferences, such as the increased reliance on \texttt{search\_temporal\_context} for M3-Bench-web and the targeted use of \texttt{listen\_to\_audio} for the audio-rich HippoVlog benchmark, demonstrating the agent's adaptive selection capabilities.

\subsection{Robustness to Graph Noise}

A well-documented limitation of base MLLMs is their tendency to hallucinate transient object interactions or behaviors, particularly in dense or occluded scenes. CAM's evidence-driven abstraction can suppress isolated errors because high-level semantics are synthesized only after a substantial history of incident edges has accumulated. To test this property directly, we inject low-level edges sampled from other graphs before abstraction and question answering. Table~\ref{tab:noise_injection} reports the resulting accuracy under increasing noise ratios.

\begin{table}[H]
    \centering
    \caption{Robustness to graph noise. We inject low-level edges sampled from other graphs before abstraction and question answering, and report accuracy (\%) on M3-Bench-robot and M3-Bench-web.}
    \label{tab:noise_injection}
    \definecolor{noiseblue}{RGB}{232, 242, 252}
    \newcommand{\noisecb}[1]{{\setlength{\fboxsep}{2pt}\colorbox{noiseblue}{\makebox[2em]{#1}}}}
    \setlength{\tabcolsep}{3pt}
    \resizebox{\linewidth}{!}{%
    \begin{tabular}{@{} l ccccc c ccccc c @{}}
        \toprule
        \multirow{2}{*}{\textbf{Injected noise ratio}} & \multicolumn{6}{c}{\textbf{M3-Bench-robot}} & \multicolumn{6}{c}{\textbf{M3-Bench-web}} \\
        \cmidrule(lr){2-7} \cmidrule(lr){8-13}
        & ME & MH & CM & PU & GK & \textbf{All} & ME & MH & CM & PU & GK & \textbf{All} \\
        \midrule
        $0\%$ (clean) & \textbf{65.5} & \textbf{80.8} & \textbf{64.8} & 82.3 & \textbf{35.3} & \noisecb{\textbf{64.5}} & \textbf{64.6} & 54.7 & \textbf{52.1} & \textbf{75.6} & 75.5 & \noisecb{\textbf{69.3}} \\
        $2\%$ & 62.4 & 77.6 & 62.6 & \textbf{83.3} & 32.3 & \noisecb{62.5} & 64.4 & \textbf{55.1} & 49.3 & 74.7 & 75.5 & \noisecb{68.5} \\
        $5\%$ & 59.4 & 79.9 & 62.4 & 82.4 & 33.8 & \noisecb{62.5} & 63.9 & 54.3 & 47.7 & 72.8 & \textbf{77.5} & \noisecb{67.9} \\
        $10\%$ & 51.5 & 73.3 & 61.4 & 81.9 & 30.4 & \noisecb{60.8} & 61.9 & 50.7 & 44.2 & 68.9 & 71.6 & \noisecb{65.2} \\
        \bottomrule
    \end{tabular}
    }
\end{table}

\subsection{Streaming and Reasoning Efficiency}

\stitle{Streaming Construction} Conditioning the extraction of clip $c_i$ on the descriptors $A_{i-1}$ from the preceding clip introduces a sequential dependency. However, this design fundamentally aligns with the constraints of our target domain: continuous, entity-centric video streams (e.g., live robotic operations, egocentric wearable cameras). In these online environments, future frames are fundamentally unavailable, rendering offline batch processing impossible. Therefore, the architectural priority is not parallelization, but rather real-time execution speed. Our empirical measurements demonstrate that processing a 30-second clip requires only 21.7 seconds on average (combined MLLM and LLM inference). Because the processing time is strictly less than the clip duration, CAM achieves faster-than-real-time performance, allowing it to scale continuously on a live video stream without inducing a temporal bottleneck.

\stitle{Reasoning Latency} Table~\ref{tab:reasoning_latency} compares average reasoning latency per question with accuracy on M3-Bench-robot.

\begin{table}[H]
    \centering
    \caption{Average reasoning latency per question and accuracy on M3-Bench-robot.}
    \label{tab:reasoning_latency}
    \setlength{\tabcolsep}{12pt}
    \begin{tabular}{@{} l cc @{}}
        \toprule
        \textbf{Method} & \textbf{Latency per question} & \textbf{Accuracy (\%)} \\
        \midrule
        M3-Agent & 71.4 s & 41.4 \\
        WorldMM & 30.0 s & 34.5 \\
        CAM (top-$k=30$) & 33.8 s & 62.9 \\
        CAM-full & 47.3 s & \textbf{64.5} \\
        \bottomrule
    \end{tabular}
\end{table}

CAM-full is 24.1 seconds faster per question than M3-Agent while improving accuracy by 23.1 percentage points. Reducing the retrieval budget to top-$k=30$ lowers latency to 33.8 seconds, close to WorldMM's 30.0 seconds, while retaining 62.9\% accuracy. This is only 1.6 percentage points below CAM-full and 28.4 points above WorldMM, demonstrating a favorable and adjustable accuracy--latency trade-off.

\subsection{Graph Visualization}

\begin{figure}[H]
    \centering
    \includegraphics[width=1\linewidth]{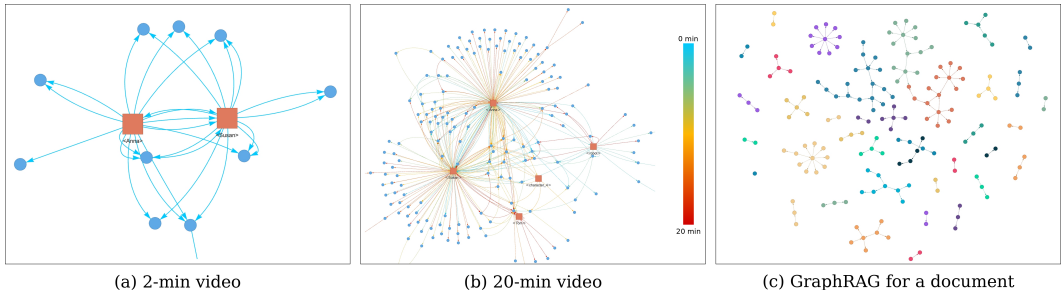}
    \caption{Qualitative comparison of graph topologies. (a) CAM memory graph for a 2-minute video. (b) CAM memory graph for a 20-minute video, demonstrating scalable and organized relational density over extended temporal horizons. (c) A standard text-based GraphRAG topology for a written document, highlighting the structural differences between general text abstraction and CAM's entity-centric video memory.}
    \label{fig:graph_topology_comparison}
\end{figure}

Long-video graphs and text-derived graphs exhibit fundamentally different topologies. Text-derived graphs typically contain many distinct nodes connected by relatively few edges, allowing densely connected communities to form meaningful clusters. In contrast, long-video graphs repeatedly accumulate observations around a small set of persistent entities, resulting in fewer nodes but substantially more temporally grounded edges. Their node-degree distribution is also highly imbalanced: principal characters accumulate far more incident evidence than secondary characters. Consequently, long-video graphs do not decompose cleanly into clusters. As illustrated in Figure~\ref{fig:graph_topology_comparison}, these properties make cluster-based abstraction poorly aligned with long-video memory. This motivates CAM's \emph{evidence-driven abstraction}, which triggers character- and relationship-level summaries according to the evidence accumulated by individual entities and entity pairs rather than relying on global graph clusters.

\section{Prompt Templates}

\newtcolorbox{promptbox}[1][]{
    enhanced,
    breakable,
    colback=black!1,
    colframe=black!35,
    colbacktitle=black!7,
    boxrule=0.45pt,
    arc=1.5pt,
    boxsep=0pt,
    left=7pt, right=7pt, top=6pt, bottom=6pt,
    lefttitle=7pt, righttitle=7pt,
    toptitle=4pt, bottomtitle=4pt,
    titlerule=0.45pt,
    before skip=10pt plus 2pt,
    after skip=10pt plus 2pt,
    fonttitle=\bfseries,
    coltitle=black,
    before upper={
        \setlist[itemize,1]{leftmargin=1.4em}
        \setlist[itemize,2]{leftmargin=1.3em}
        \setlist[enumerate,1]{leftmargin=1.7em}
        \setlist[enumerate,2]{leftmargin=1.5em}
    },
    title=#1
}

In this section, we provide the specific prompts used to instruct the LLM planner and extract abstract information within the CAM framework.

\subsection{Memorization Prompts}

\begin{promptbox}[Prompt: Clip-level Observation]
You are given a 30-second video represented as sequential frames (pictures in chronological order).

\textbf{Your tasks:}
\begin{enumerate}
    \item \textbf{Characters' Behavior}: Describe each character's behavior in chronological order.
    \begin{itemize}
        \item Include: (a) Interaction with objects in the scene. (b) Interaction with other characters. (c) Actions and movements.
        \item Do NOT repeat the information from the conversation.
        \item When the character interacts with objects in the scene, include precise location information for placement, retrieval, or movement.
        \item Furniture/container names include "dressing table", "bedside table", "wardrobe", "shoe cabinet", "refrigerator", and "microwave oven".
        \item Spatial modifiers include "below", "above", "left side", "right side", "beside", "in front of", "next to", "under", and "on the counter".
        \item Complete location descriptions: Always combine furniture names with spatial modifiers to form hierarchical locations (e.g., "cabinet below the dressing table", "second layer of the refrigerator", "cabinet on the left side of the wardrobe"). For retrieval actions, include source locations (e.g., "takes towel from Susan's bag", "gets mask from cabinet below dressing table").
        \item Each entry must describe exactly one event/detail. Split sentences if needed.
    \end{itemize}
    \textit{Example output:} \texttt{["<Alice> enters the room.", "<Alice> takes cap from the cabinet on the left side of the wardrobe.", "<Alice> sits with <Bob> side by side on the couch.", "<Bob> watches TV."]}

    \item \textbf{Conversation}: Record the dialogue based on subtitles. Always use the characters' real name that appears in the subtitle.
    \begin{itemize}
        \item \textit{Output format:} List of two-element lists \texttt{[character, content]}.
        \item \textit{Example output:} \texttt{[["<Alice>", "Hello, my name is Alice."], ["<Bob>", "Hi, I'm Bob. Nice to meet you."]]}
    \end{itemize}

    \item \textbf{Characters' Appearance}: Describe or update each character's appearance, including facial features, clothing, body shape, hairstyle, or other distinctive characteristics.
    \begin{itemize}
        \item Each characteristic should be concise, separated by commas.
        \item Existing characters: Update if changes observed (hair, clothing), enhance if new details visible, otherwise keep unchanged.
        \item Keep appearance info even if character leaves scene. If a character is not visible in scene, do not include them in the characters' appearance list.
    \end{itemize}
    \textit{Example output:} \texttt{[Appearance(name="<Alice>", appearance="female, fat, ponytail, wear glasses, short-sleeved shirt, blue jeans, white sneakers")]}

    \item \textbf{Scene}: Use one word or phrase to describe the scene in the current video (e.g., "bedroom", "gym", "office"). This field is optional and should only be provided if the scene is clearly identifiable.

\end{enumerate}

\vspace{0.5em}
\textbf{Character Naming Rules}:\\
Use angle brackets to represent characters (e.g., <Alice>, <Bob>, <male\_1>) in behaviors, conversation, and character appearance. There are three cases to name a character:
\begin{enumerate}
    \item The name can be deduced from the conversation or subtitles. Refer to the character by his/her name (e.g., <Alice>, <Bob>).
    \item The name is not provided, but you can deduce the character's job or identity (e.g., police, mailsman). Refer to the character by \texttt{<job\_number>} (e.g., <police\_1>). Only refer to a character by their job when it is clear. Do not make any unsure guesses on a character's job.
    \item Neither the name nor the job can be inferred. Name the character by gender and a number (e.g., <male\_1>, <male\_2>, <female\_1>).
\end{enumerate}
For the second and third cases, if the character's name is provided in the following video clips, use the equivalence line to update the character's name.

\vspace{0.5em}
\textbf{Character Matching Rules}:\\
For characters appearing in the video (MUST follow before creating new characters):
\begin{itemize}
    \item First, check if the subtitle name is provided. If so, use it as the character name (e.g., <Anna>). Also compare the character name with the unknown characters (e.g., <male\_1>, <police\_1>) in the appearance list. If high similarity is found, add "Equivalence: <unknown\_id>, <Anna>" at the start of behaviors. This indicates that the character is the same as the previously identified unknown character. If the equivalence is found, refer to this character by its character name instead of the unknown ID in the behaviors, conversation, and character appearance.
    \item If a character's name is not provided in the subtitle, match this character's appearance to the characters in the appearance list. If high similarity is found, refer to this character by its name in the appearance list.
    \item If no match is found, create a new character following the naming rules above. Our goal is to minimize the number of unique character IDs. When uncertain, match to existing rather than creating new.
\end{itemize}

\vspace{0.5em}
\textbf{Additional Rules}:
\begin{itemize}
    \item Maintain strict chronological order.
    \item Avoid repetition in both behavior and conversation. If no behavior or conversation is observed, return an empty list for behaviors and conversation.
\end{itemize}
\end{promptbox}

\begin{promptbox}[Prompt: Triples Extraction]

Convert each action sentence into triples of the form
\texttt{[source, content, target]}.

Return \textbf{only} a valid JSON array. Do not include explanations, Markdown, or any extra text.

\textbf{Output Format:}
\begin{itemize}[leftmargin=1.4em]
    \item Use the format \texttt{[[source, content, target], ...]} with double quotes and no trailing commas.
    \item Use \texttt{null} when no target exists.
    \item Preserve the original sentence order and the action order within each sentence.
\end{itemize}

\textbf{Definitions:}
\begin{itemize}[leftmargin=1.4em]
    \item \textbf{Source}: The entity performing the action or having the state.
    \item \textbf{Content}: A verb-centered action, relation, or state.
    \item \textbf{Target}: The affected or related entity.
\end{itemize}

First identify actors, actions, and affected entities; then resolve references, split compound structures, normalize verbs, add state relations, and remove redundancy.

\textbf{Rules:}
\begin{enumerate}[leftmargin=1.7em]
    \item \textbf{Entities}: Keep angle brackets around characters (e.g., \texttt{<Alice>}, \texttt{<robot>}) but never around objects. If an object has angle brackets in the input, remove them. Copy entity names verbatim and do not invent entities.

    \item \textbf{Verbs}: Use simple present tense rather than progressive forms. Retain relevant prepositions, directions, and adverbs (e.g., ``looks at,'' ``turns left,'' or ``runs quickly'').

    \item \textbf{Objects}: Use singular nouns, retain adjectives and named-object descriptions, and split compound objects into separate triples.

    \item \textbf{Multiple Relations}: Create a separate triple for each subject, action, and object while preserving their order. For example, ``\texttt{<Lisa>} dances and sings'' produces two triples.

    \item \textbf{Reference Resolution}: Never retain pronouns such as ``his,'' ``her,'' or ``their.'' Replace them with explicit entities or owners; if ambiguous, use the nearest subject.

    \item \textbf{Body Parts}: Merge a body part into the action instead of treating it as an object. For example, ``\texttt{<Alice>} hits \texttt{<Bob>}'s head'' becomes \texttt{["<Alice>", "hits head", "<Bob>"]}.

    \item \textbf{Communication}: Encode communication directly between characters. Do not create abstract targets such as ``question'' or ``message''; for example, use \texttt{["<Tom>", "asks", "<Mary>"]}.

    \item \textbf{Resulting States}: Add a state triple when an action changes an object's state or location. For example, ``puts coffee on table'' yields \texttt{["<robot>", "puts", "coffee"]} and \texttt{["coffee", "is on", "table"]}.

    \item \textbf{Spatial Relations}: Keep location information in the content field and use clean entities as the source and target. For example, use \texttt{["<Betty>", "sits on the right side of", "sofa"]}.

    \item \textbf{Consistency}: Preserve meaningful adjectives, adverbs, and other modifiers. Keep only distinct actions and do not duplicate a state already expressed by a stronger relation.

    \item \textbf{Fallback}: If unsure, use the minimal form \texttt{[source, verb, target]}.
\end{enumerate}

\textbf{Example Input:}\\
\texttt{["<Michael> pats <Susan>'s shoulder and smiles.",}\\
\texttt{"<robot> places the red cup on the counter.",}\\
\texttt{"<John> takes his wallet and keys from the drawer.",}\\
\texttt{"<Betty> carries a red bag in her left hand and a white bag in her right hand."]}

\textbf{Example Output:}\\
\texttt{[["<Michael>", "pats shoulder", "<Susan>"],}\\
\texttt{["<Michael>", "smiles", null],}\\
\texttt{["<robot>", "places", "red cup"],}\\
\texttt{["red cup", "is on", "counter"],}\\
\texttt{["<John>", "takes", "John's wallet"],}\\
\texttt{["<John>", "takes", "John's key"],}\\
\texttt{["John's wallet", "is in", "drawer"],}\\
\texttt{["John's key", "is in", "drawer"],}\\
\texttt{["<Betty>", "carries in left hand", "red bag"],}\\
\texttt{["<Betty>", "carries in right hand", "white bag"]]}

Now convert the following list of action sentences into triples:

\end{promptbox}

\begin{promptbox}[Prompt: Character-level Abstraction]

You are given a character's name and a list of their behaviors in chronological order.

Your task is to summarize the character's attributes:
\begin{itemize}
    \item Personality (e.g., confident or nervous).
    \item Role or profession (e.g., host or newcomer).
    \item Interests or background, when inferable.
    \item Distinctive behaviors or traits (e.g., speaks formally or fidgets).
\end{itemize}
Avoid restating visual facts; focus on identity construction.

For each attribute, you should also provide a confidence score between 0 and 100.
If the confidence score is less than 50, you should not include the attribute in the output.

Output a JSON dictionary in which each key is an attribute and each value is its confidence score.

\textit{Example output:} \texttt{\{"student": 90, "enthusiastic": 80, "likes to read": 70, "professional": 50, "likes to play games": 60\}}

\end{promptbox}

\begin{promptbox}[Prompt: Relationship-level Abstraction]

You are given a list of character interactions in chronological order.
Your task is to extract relationships between the characters, including:
\begin{itemize}
    \item Roles (e.g., friends, colleagues, host--guest, teacher--student, or parent--child).
    \item Attitudes or emotions (e.g., respect, dislike, or friendliness).
    \item Power dynamics (e.g., who leads or whether the characters are equal).
    \item Evidence of cooperation.
    \item Exclusion, conflict, or competition.
\end{itemize}

\textbf{Additional Rules:}
\begin{itemize}
    \item Store only abstract relationships between the characters.
    \item Do not include actions or summaries of actions in the output (e.g., \texttt{<Alice>} speaks with \texttt{<Bob>} or \texttt{<Alice>} plays games with \texttt{<Bob>}).
    \item Do not generate repetitive or symmetric information.
\end{itemize}

For each relationship, you should also provide a confidence score between 0 and 100.
If the confidence score is less than 50, you should not include the relationship in the output.
It is acceptable to only generate a few relationships if you don't have enough information.

Output a JSON array (a list of lists). Each inner list contains four elements: \texttt{[character1, relationship, character2, confidence score]}.

\textit{Example output:}\\
\texttt{[["<Alice>", "is friend with", "<Bob>", 90],}\\
\texttt{["<Alice>", "is teacher of", "<Charlie>", 80],}\\
\texttt{["<Charlie>", "respects", "<Alice>", 70]]}

\end{promptbox}

\begin{promptbox}[Prompt: Conversation Summary]

You are given a conversation between several characters.

\textbf{Output Format:}
Return a JSON object with the following keys:
\begin{enumerate}
    \item \texttt{"summary"}: A string summarizing the key topics, decisions, or outcomes in two to four concise sentences.
    \item \texttt{"character\_attributes"}: A list of \texttt{[character, attribute, confidence\_score]} triplets.
    \item \texttt{"characters\_relationships"}: A list of \texttt{[character1, relationship, character2, confidence\_score]} quadruplets.
\end{enumerate}

\textit{Example output:}\\
\texttt{\{"summary": "Alice and Bob discussed their upcoming project.}\\
\texttt{They agreed on a timeline and assigned tasks.",}\\
\texttt{"character\_attributes": [["<Alice>", "organized", 85],}\\
\texttt{["<Bob>", "cautious", 70]],}\\
\texttt{"characters\_relationships":}\\
\texttt{[["<Alice>", "is friend with", "<Bob>", 90]]\}}

\vspace{0.5em}
\textbf{Detailed Instructions:}
\begin{enumerate}
    \item \textbf{Summary}
    \begin{itemize}
        \item Summarize the key topics, decisions, or outcomes discussed in the conversation.
        \item Write two to four concise sentences covering the main themes and important points.
        \item Focus on what was discussed and decided, rather than on individual statements.
    \end{itemize}
    \textit{Example:} ``Alice and Bob discussed their upcoming project. They agreed on a timeline and assigned tasks. Bob expressed concerns about the deadline, which Alice addressed by suggesting additional resources.''

    \item \textbf{Character Attributes}
    \begin{itemize}
        \item Extract each character's attributes revealed through their dialogue and interaction style.
        \item Focus on personality traits, role or profession, interests, and background information when mentioned.
        \item Do not include physical appearance, concrete actions, temporary emotional states, or information not directly supported by the conversation.
        \item Confidence scores range from 0 to 100. Include only attributes with confidence scores of at least 50.
        \item Avoid redundant or overly similar attributes (e.g., do not include both ``friendly'' and ``kind'' unless they are distinctly different).
        \item Use angle brackets for character names (e.g., \texttt{<Alice>} and \texttt{<Bob>}).
        \item Output a list of three-element lists: \texttt{[character, attribute, confidence\_score]}.
    \end{itemize}
    \textit{Example output:}\\
    \texttt{[["<Alice>", "organized", 85], ["<Alice>", "problem-solver", 75],}\\
    \texttt{["<Bob>", "detail-oriented", 80], ["<Bob>", "cautious", 70]]}\\
    \textit{Invalid examples:} \texttt{["<Alice>", "asked a question", 90]} describes an action, and \texttt{["<Bob>", "has brown hair", 80]} describes appearance.

    \item \textbf{Character Relationships}
    \begin{itemize}
        \item Extract abstract relationships between characters based on their dialogue interactions.
        \item Include roles, attitudes, power dynamics, and evidence of cooperation, conflict, exclusion, or competition.
        \item Do not include specific actions or events, temporary interactions, dialogue content, or discussion topics.
        \item Confidence scores range from 0 to 100. Include only relationships with confidence scores of at least 50.
        \item Do not generate symmetric duplicates. If \texttt{<Alice>} respects \texttt{<Bob>}, do not include the reverse unless it is explicitly different.
        \item It is acceptable to generate only a few relationships if there is insufficient information.
        \item Output a list of four-element lists: \texttt{[character1, relationship, character2, confidence\_score]}.
    \end{itemize}
    \textit{Example output:}\\
    \texttt{[["<Alice>", "is friend with", "<Bob>", 90],}\\
    \texttt{["<Alice>", "is teacher of", "<Charlie>", 80],}\\
    \texttt{["<Charlie>", "respects", "<Alice>", 70]]}\\
    \textit{Invalid examples:} \texttt{["<Alice>", "spoke with", "<Bob>", 90]} is an action, and \texttt{["<Alice>", "discussed the project", "<Bob>", 85]} is dialogue content.
\end{enumerate}

\textbf{Edge Cases:}
\begin{itemize}
    \item If only one character speaks, focus on that character's attributes and skip relationships.
    \item If the conversation is empty or unclear, return empty arrays for attributes and relationships and provide a brief summary noting the issue.
    \item If character names are ambiguous, use the names provided in the conversation.
\end{itemize}

Now summarize the following conversation in JSON format:

\end{promptbox}

\subsection{Reasoning Prompts}

\begin{promptbox} [Prompt: Planner Strategy]
\textbf{Tool usage guidance:}
\begin{itemize}
    \item \texttt{general\_search}: Use this FIRST for all questions to get a temporal anchor and find relevant \texttt{clip\_id}s. Allocate your budget (Total $k \le 50$) based on the primary modality of the question:
    \begin{itemize}
        \item \textbf{k\_action (0-30)}: Primary for behavior, actions, temporal sequence, or `where is' queries.
        \item \textbf{k\_conversation (0-30)}: Primary for `why', dialogue, or causal reasoning.
        \item \textbf{k\_ocr (0-30)}: Primary for text on signs, labels, or posters.
        \item \textbf{k\_high\_level (0-10)}: Secondary for character traits or relationships.
        \item \textbf{k\_appearance (0-15)}: Use ONLY for physical looks, hair, or clothing. Set to 0 if the question is irrelevant to appearance.
    \end{itemize}
    \textbf{IMPORTANT (First Round)}: If this is your first tool call, you MUST use \texttt{general\_search} and allocate the FULL budget (Total $k=50$) across these modalities to ensure a broad understanding of the video.

    \item \texttt{search\_temporal\_context}: Use this ONLY after finding a candidate \texttt{clip\_id} via \texttt{general\_search} to see events right before or after it.

    \item \texttt{watch\_video\_clip}: Use this ONLY after finding a candidate \texttt{clip\_id} via \texttt{general\_search}. Mandatory for visual questions requiring high detail (e.g., specific placement of objects, visual state, exact counts) when text graph is insufficient. Provide a specific \texttt{focus} based on what is missing from the text.
\end{itemize}

\textbf{Strategy for Clip ID Selection:}
\begin{enumerate}
    \item Look for the \texttt{[clip\_id]} next to the most relevant action or conversation in \texttt{general\_search} results.
    \item Pay attention to temporal context: if looking for what happened \textit{before} an event, search the clip(s) preceding the event.
    \item Use that specific \texttt{clip\_id} as input for \texttt{watch\_video\_clip} or \texttt{search\_temporal\_context}.
\end{enumerate}

Analyze the conversation history above. What is the most effective next step to solve the question?
\end{promptbox}

\begin{promptbox}[Prompt: Query Parser]
You are a query parser for a knowledge graph system that stores video information in a hierarchical structure.

Given a query, output the following in JSON format:

\textbf{Output Fields:}
\begin{enumerate}
    \item \texttt{query\_triples}: A list containing one to three query triples.
    \begin{itemize}
        \item Each triple has the format \texttt{[source, content, target, source\_weight, content\_weight, target\_weight]}.
        \item Use \texttt{null} for missing components.
        \item Normalize unnamed characters to generic IDs (e.g., \texttt{<male\_1>}); retain a name when it is explicitly provided.
        \item Assign each weight a value between 0.0 and 1.0.
    \end{itemize}

    \item \texttt{speaker\_strict}: Set this field to the relevant character ID or IDs when the query asks about specific dialogue; otherwise, use \texttt{null}.

    \item \texttt{spatial\_constraint}: Use a location string only when the query refers to a general space; otherwise, use \texttt{null}.
\end{enumerate}

\textbf{Special Rules for Location Queries:}
\begin{itemize}
    \item \textbf{Hierarchical locations}: Preserve complete location phrases (e.g., ``cabinet below the dressing table'') without splitting them.
    \item \textbf{Temporal-spatial queries}: Construct triples that retrieve the most recent state or source location.
\end{itemize}

Now parse the following query in JSON format:
\end{promptbox}

\begin{promptbox}[Prompt: Final Answer]
You are the final-answer synthesizer. Use all collected evidence in the conversation history to answer the original question concisely and directly.

\textbf{Requirements:}
\begin{itemize}
    \item Respond in exactly \textbf{one sentence}.
    \item Do not include explanations, meta-commentary, or justifications.
    \item Do not answer ``I don't know,'' ``The information is not sufficient,'' or ``It is unclear.''
    \item If the evidence is uncertain, make the most reasonable inference supported by the conversation history.
    \item Reuse exact terms from the question and search results when possible.
\end{itemize}
\end{promptbox}

\subsection{Answer Verification}
\label{sec:answer_verification}
\begin{promptbox}[Prompt: Answer Verifier (M3-Bench Only)]
You are provided with a question, a ground truth answer, and an answer from an agent model. Your task is to determine whether the ground truth answer can be logically inferred from the agent's answer, in the context of the question.

Do not compare only the surface forms of the two answers. Instead, determine whether the meaning of the agent answer supports or implies the ground truth answer in the context of the question.

\textbf{Decision Rule:}
\begin{itemize}
    \item Return \texttt{"Yes"} if the ground truth answer can be reasonably inferred from the agent answer.
    \item Return \texttt{"No"} otherwise.
\end{itemize}

\textbf{Important Notes:}
\begin{itemize}
    \item Exact wording or matching sentence structure is not required.
    \item Semantic entailment or implication is sufficient.
    \item Return only \texttt{"Yes"} or \texttt{"No"}, with no explanation or additional formatting.
\end{itemize}

\textbf{Input Fields:}
\begin{itemize}
    \item \texttt{question}: The question asked.
    \item \texttt{ground\_truth\_answer}: The correct answer.
    \item \texttt{agent\_answer}: The model answer to be evaluated.
\end{itemize}

Now evaluate the following input:

\textbf{Input:}\\
\texttt{question: \{question\}}\\
\texttt{ground\_truth\_answer: \{ground\_truth\_answer\}}\\
\texttt{agent\_answer: \{agent\_answer\}}

\textbf{Output:} \texttt{"Yes"} or \texttt{"No"}
\end{promptbox}

HippoVlog does not require an LLM-as-a-judge because its multiple-choice answers can be verified directly.



\newpage

\end{document}